\documentclass[conference]{IEEEtran}
\IEEEoverridecommandlockouts
\usepackage{cite}
\usepackage{amsmath,amssymb,amsfonts}
\usepackage{algorithmic}
\usepackage{graphicx}
\usepackage{textcomp}
\usepackage{xcolor}

\usepackage[utf8]{inputenc} 
\usepackage[T1]{fontenc}    
\usepackage{hyperref}       
\usepackage{url}            
\usepackage{booktabs}       
\usepackage{amsfonts}       
\usepackage{nicefrac}       
\usepackage{microtype}      
\usepackage{lipsum}
\usepackage{graphicx}

\usepackage{multirow}
\usepackage{algorithm}
\usepackage{algorithmic}

\usepackage{caption}
\usepackage{subcaption}
\usepackage{float}

\usepackage{bbm}
\usepackage{multicol}

\def\BibTeX{{\rm B\kern-.05em{\sc i\kern-.025em b}\kern-.08em
    T\kern-.1667em\lower.7ex\hbox{E}\kern-.125emX}}
\begin{document}

\title{
Linguistic Laws Meet Protein Sequences: A Comparative Analysis of Subword Tokenization Methods
}

\author{\IEEEauthorblockN{Burak Suyunu}
\IEEEauthorblockA{\textit{Computer Engineering} \\
\textit{Boğaziçi University}\\
İstanbul, Türkiye\\
burak.suyunu@std.bogazici.edu.tr}
\and
\IEEEauthorblockN{Enes Taylan}
\IEEEauthorblockA{\textit{Computer Engineering} \\
\textit{Boğaziçi University}\\
İstanbul, Türkiye\\
enes.taylan1@std.bogazici.edu.tr}
\and
\IEEEauthorblockN{Arzucan Özgür}
\IEEEauthorblockA{\textit{Computer Engineering} \\
\textit{Boğaziçi University}\\
İstanbul, Türkiye\\
arzucan.ozgur@bogazici.edu.tr}
}

\maketitle

\begin{abstract}
Tokenization is a crucial step in processing protein sequences for machine learning models, as proteins are complex sequences of amino acids that require meaningful segmentation to capture their functional and structural properties.
However, existing subword tokenization methods, developed primarily for human language, may be inadequate for protein sequences, which have unique patterns and constraints.
This study evaluates three prominent tokenization approaches, Byte-Pair Encoding (BPE), WordPiece, and SentencePiece, across varying vocabulary sizes (400–6400), analyzing their effectiveness in protein sequence representation, domain boundary preservation, and adherence to established linguistic laws.
Our comprehensive analysis reveals distinct behavioral patterns among these tokenizers, with vocabulary size significantly influencing their performance. BPE demonstrates better contextual specialization and marginally better domain boundary preservation at smaller vocabularies, while SentencePiece achieves better encoding efficiency, leading to lower fertility scores. WordPiece offers a balanced compromise between these characteristics. However, all tokenizers show limitations in maintaining protein domain integrity, particularly as vocabulary size increases.
Analysis of linguistic law adherence shows partial compliance with Zipf's and Brevity laws but notable deviations from Menzerath's law, suggesting that protein sequences may follow distinct organizational principles from natural languages.
These findings highlight the limitations of applying traditional NLP tokenization methods to protein sequences and emphasize the need for developing specialized tokenization strategies that better account for the unique characteristics of proteins. Our work contributes to the ongoing dialogue between bioinformatics and natural language processing, offering insights for future development of protein-specific tokenization approaches.
\end{abstract}


\section{Introduction}
\label{section:introduction}
In recent years, bioinformatics has increasingly adopted natural language processing (NLP) techniques to analyze and model protein sequences.
Since proteins can be represented as sequences of amino acids, there is increasing interest in treating them as a type of biological language, enabling applications such as protein function prediction, structural analysis, and interaction modeling~\cite{nambiar2020transforming, rives2021biological, elnaggar2021prottrans, ofer2021language, brandes2022proteinbert, lin2023evolutionary, elnaggar2023ankh}.

Tokenization is a critical step in this process, as it defines how sequences are segmented, directly influencing model performance and interpretability.
However, proteins have distinct characteristics from natural languages, such as complex long-range dependencies and structural properties that traditional NLP tokenization approaches may not capture.
These differences raise essential questions about the effectiveness of standard NLP tokenization methods when applied to protein sequences.

Recent studies have begun to explore these questions. Tan et al.~\cite{tan2023peta} and Dotan et al.~\cite{dotan2024effect} investigated how tokenization methods and vocabulary sizes affect the performance of protein language models. Their work reveals that vocabulary size significantly affects protein representation, with larger vocabularies often leading to degraded performance in structure prediction tasks.
Building on this foundation, Ieremie et al.~\cite{ieremie2024protein} explored how different representations of protein sequences, particularly reduced amino acid alphabets, affect model behavior and interoperability.
While these studies provide valuable insights, they primarily focus on downstream task performance rather than analyzing the fundamental properties of different tokenization approaches.

Research by Vig et al.~\cite{vig2020bertology} and Rao et al.~\cite{rao2020transformer} on the interpretability of protein language models has shown that attention patterns in transformers can capture biologically meaningful patterns, such as conserved regions and structural motifs. The importance of thoughtful tokenization strategies extends beyond protein sequences. Similar investigations in related fields, such as the analysis of SMILES representations for protein-ligand binding~\cite{temizer2024exploring}, demonstrate how careful tokenization choices can reveal meaningful biological patterns.
While these studies focus on analyzing attention patterns and model interpretability, our work takes a step back to examine the fundamental properties of the tokenization methods themselves, investigating how different approaches segment protein sequences and how these segmentation patterns relate to biological and linguistic principles.

The relationship between biological sequences and linguistic principles has also attracted attention. Shahzad et al.~\cite{shahzad2015organization} and Semple et al.~\cite{semple2022linguistic} have demonstrated that protein organization follows certain linguistic laws, such as Menzerath-Altmann's Law and Zipf's Law. These findings suggest that linguistic principles might offer valuable insights into protein structure and organization.
While recent work has explored structure-aware approaches~\cite{su2023saprot, heinzinger2023bilingual, van2024fast}, the relationship between tokenization methods and linguistic laws in protein sequences remains understudied.

Our study builds upon and extends this prior research by providing a comprehensive evaluation of three popular subword tokenization methods originally developed for natural languages, Byte-Pair Encoding (BPE)~\cite{sennrich2016neural}, WordPiece~\cite{wu2016google}, and SentencePiece~\cite{kudo2018sentencepiece}, to understand how well they capture the underlying language in protein sequences.
Unlike previous work that primarily focused on downstream task performance, we evaluate these tokenizers from multiple angles: their ability to handle vocabulary scaling, alignment with protein domain boundaries, effectiveness on tokenization-related metrics, and adherence to linguistic laws (Zipf's, Brevity, Heaps', and Menzerath's laws).

The motivation behind this comparison is not only to identify which tokenization method performs best but also to investigate how closely the language of proteins aligns with established linguistic principles. Such insights could reveal structural patterns unique to proteins, guiding future efforts to develop domain-specific tokenization methods that are better suited for proteins. Ultimately, our work contributes to a deeper understanding of the parallels and divergences between natural and biological languages, offering a foundation for improved models for protein function prediction, structure analysis, and other computational biology tasks.

\section{Subword Tokenization Methods}
\label{section:background-tokenization}

Tokenization involves dividing a text into smaller units such as words, phrases, symbols or other meaningful elements, called tokens. This step is usually the first in any text processing pipeline, and the selection of the tokenization method can significantly influence the outcome of subsequent NLP operations. Thus, it is crucial to consider the most appropriate tokenization approach for a specific task. 

Subword tokenization methods are based on the idea that commonly used words should not be broken down into smaller subwords, while infrequent words should be divided into meaningful subparts. For example, the word \emph{quietly} might be considered a rare word and split into \emph{quiet} and \emph{ly}. Both \emph{quiet} and \emph{ly} appear more frequently as standalone subwords, and at the same time, the meaning of \emph{quietly} is preserved by the combination of \emph{quiet} and \emph{ly}. Subword tokenization enables the model to process new words by breaking them into familiar subwords. The most well-known algorithms for subword tokenization are Byte-Pair Encoding (BPE) ~\cite{sennrich2016neural}, WordPiece~\cite{wu2016google}, SentencePiece~\cite{kudo2018sentencepiece}, and Unigram~\cite{kudo2018subword}. Except for SentencePiece, these subword tokenizers use a pre-tokenizer (space tokenization) to divide the training data into words. BPE starts with an initial vocabulary that includes all the symbols in the dataset. Then, it repeatedly forms new symbols by merging the two most frequent symbols until the desired vocabulary size is reached. WordPiece is similar to BPE, but instead of merging symbols based on frequency, it merges symbols based on the likelihood of the training data after the new symbol is added to the vocabulary. Unlike BPE or WordPiece, Unigram starts with many tokens and iteratively discards tokens to obtain a smaller vocabulary. The Unigram algorithm calculates a loss (often defined as log-likelihood) over the training data given the current vocabulary and a Unigram language model. SentencePiece treats the input as a raw input stream and includes the space in the set of characters to be used, and then it applies the BPE or Unigram algorithms to create an appropriate vocabulary.

\section{Dataset}
\label{section:background-dataset}

We used the UniRef50 dataset~\cite{10.1093/bioinformatics/btm098} for protein tokenizers and the WikiText corpus for the English tokenizer.
The UniRef database clusters protein sequences at various identity levels to reduce redundancy, with UniRef50 clustering sequences that share at least 50\% identity. This approach minimizes the oversampling of evolutionarily similar sequences, offering a diverse and representative set for protein sequence analysis.

We downloaded the UniRef50 dataset from HuggingFace\footnote{\url{https://huggingface.co/datasets/agemagician/uniref50}}. The subword tokenizers were trained with randomly selected 15 million sequences from the data's train split. Then, the experiments were applied to the combination of validation and test splits (11957 sequences). We discarded 14 sequences from the test set that are longer than 3k residues.

We also downloaded the English WikiText language modeling dataset \cite{merity2016pointer} from HuggingFace\footnote{\url{https://huggingface.co/datasets/Salesforce/wikitext}}. A BPE tokenizer was trained on the data's train split (4.2 million sentences). Then, the experiments were applied to the combination of validation and test splits (19720 sentences).

\section{Experiments and Analyses}
\label{section:experiments}

In our experiments, we evaluate three subword tokenization methods: Byte-Pair Encoding (BPE), WordPiece, and SentencePiece with the Unigram language model, on protein sequences using the UniRef50 dataset. 
To compare these methods across different granularities, we trained each tokenizer at varying vocabulary sizes: 400, 800, 1600, 3200, and 6400.
Additionally, we trained a BPE tokenizer using the English WikiText corpus to provide a natural language baseline, allowing us to explore differences in tokenization between natural language and protein sequences.

We assessed each tokenizer based on a range of metrics that reflect critical aspects of sequence segmentation and representation.
These metrics include shared token percentages, token length distribution, fertility, and contextual exponence, which together indicate how each tokenizer scales, maintains consistency across data, and segments sequences.
To further examine each method’s capability, we evaluated their alignment with protein domain boundaries and their adherence to statistical patterns observed in linguistic data, such as Zipf’s, Brevity, Heaps’, and Menzerath’s laws.
This analysis aims to uncover the underlying structures within protein sequences, shedding light on how different tokenization methods might capture patterns beyond surface-level segmentation.

\subsection{Shared Token Percentages}
\label{section:shared-token-percentages}

\begin{figure}[htbp]
	\begin{center}
		\includegraphics[width=.85\columnwidth]{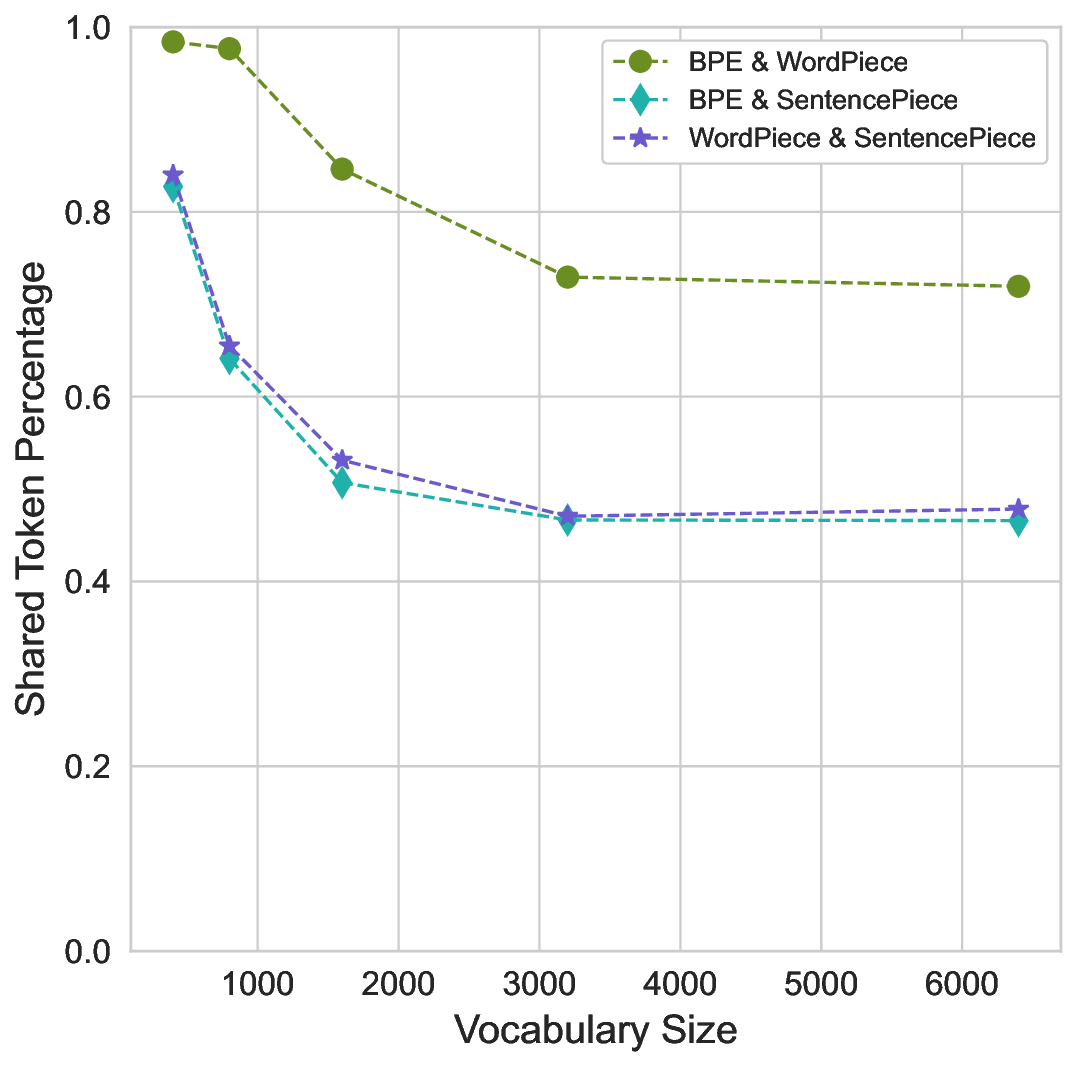}
		\caption{The plot of percentage of shared tokens between different pairs of tokenizers across different vocabulary sizes.}
		\label{fig:shared_2}
	\end{center}
\end{figure}

Analysis of shared token percentages reveals interesting patterns across vocabulary sizes, as seen in Fig.~\ref{fig:shared_2}. With small vocabularies of 400 tokens, BPE and WordPiece show high overlap (0.98), while both maintain moderate overlap with SentencePiece (0.83-0.84), suggesting tokenizer choice is less critical at this scale. As vocabularies grow to 6400 tokens, overlap decreases between BPE and WordPiece (0.72), with SentencePiece diverging significantly (0.47 overlap with both), indicating increasing uniqueness in each method's vocabulary. This consistent trend of decreasing overlap with increasing vocabulary size highlights each tokenizer's distinct approach becoming more apparent at larger scales. SentencePiece consistently shows the least overlap, reflecting its fundamentally different strategy in token generation. These findings reveal the growing importance of tokenizer selection as vocabulary size increases.

\subsection{Token Length Distribution and Fertility}

\begin{figure}[htbp]
    \begin{center}
    \subfloat[]{\includegraphics[width = .48\columnwidth]{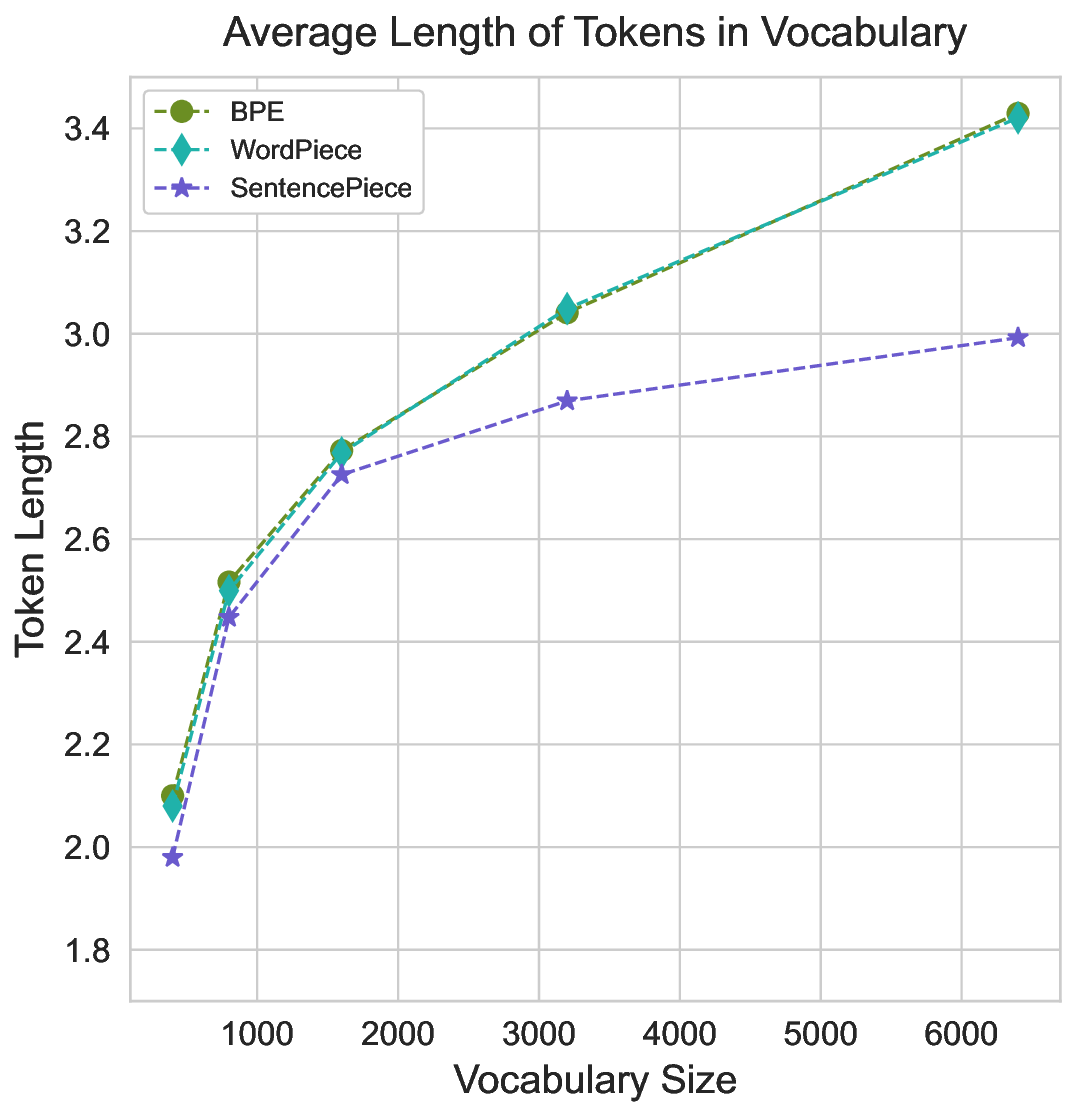}\label{subfig:token_len_fertility_a}}\hfill
    \subfloat[]{\includegraphics[width = .48\columnwidth]{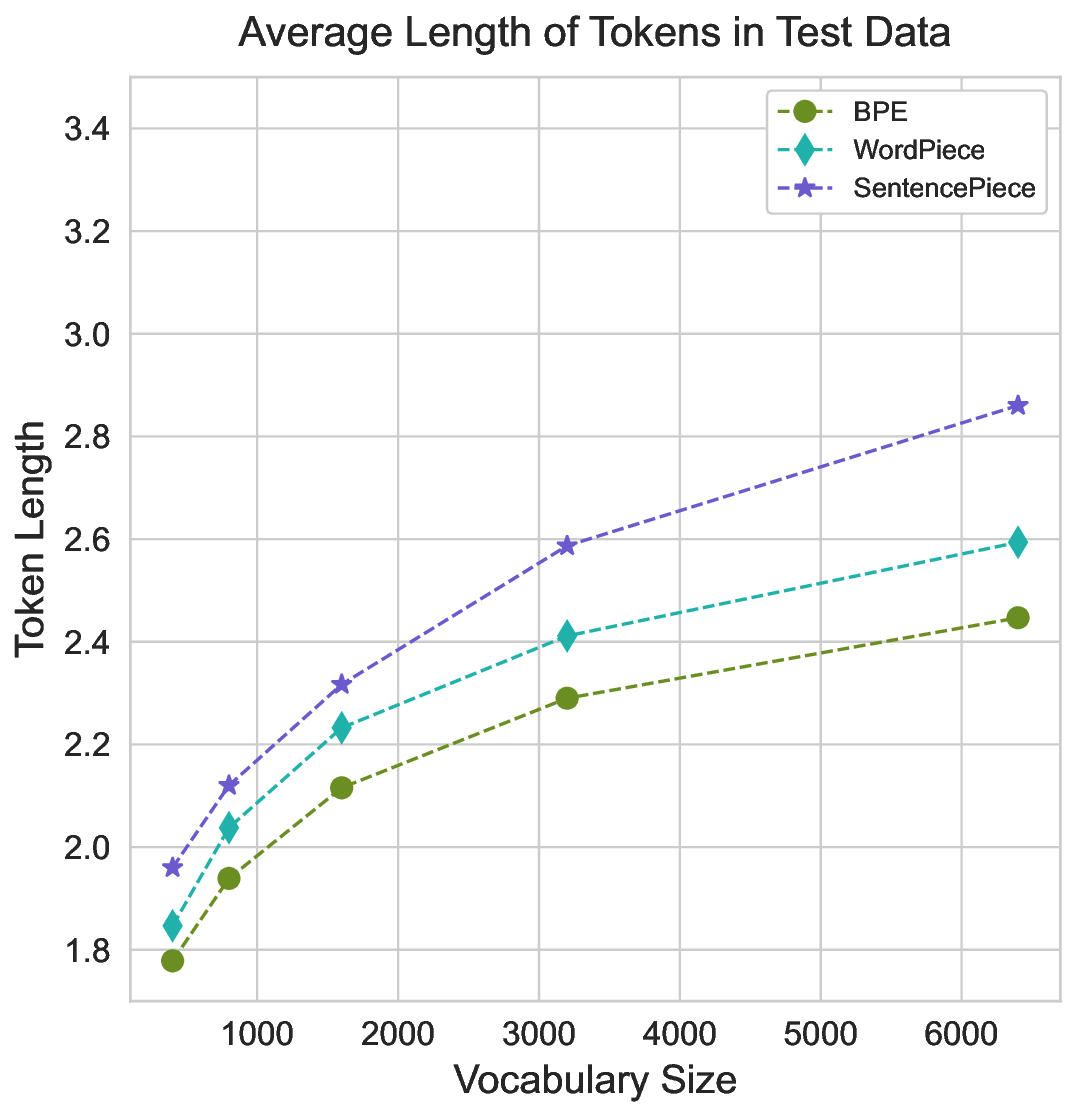}\label{subfig:token_len_fertility_a2}}
    
    \subfloat[]{\includegraphics[width = .48\columnwidth]{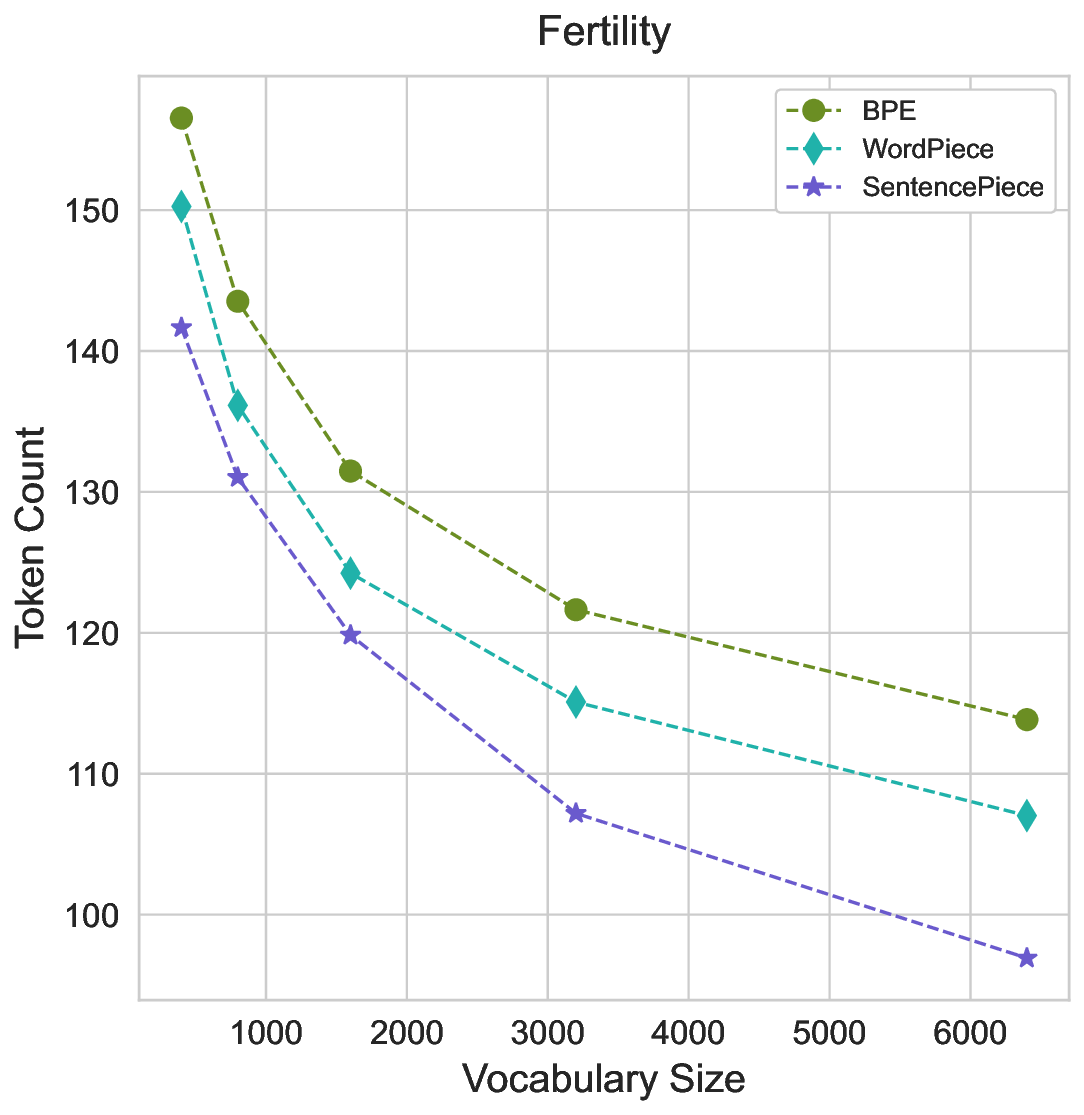}\label{subfig:token_len_fertility_b}}
    \caption{The plots of average lengths of tokens in (a) vocabulary and (b) test data and (c) fertility scores of BPE, WordPiece, and SentencePiece across different vocabulary sizes.}
    \label{fig:token_len_fertility}
    \end{center}
\end{figure}

In this section, we focus on token length distribution and fertility, offering insights beyond the basic relationship between vocabulary size and token length or token count.
As shown in Fig.~\ref{subfig:token_len_fertility_a}, BPE consistently produces the longest average token length in the vocabulary learned from the training set for all tested vocabulary sizes, closely followed by WordPiece, while SentencePiece yields shorter tokens on average. However, this trend is reversed when we examine the average token length in the test data, as seen in Fig.~\ref{subfig:token_len_fertility_a2}. Here, BPE produces the shortest tokens on average, followed by WordPiece, while SentencePiece generates the longest.
This discrepancy in token length across training and test data directly impacts fertility, as shown in Fig.~\ref{subfig:token_len_fertility_b}. Fertility refers to the number of tokens needed to encode a sequence.
BPE's shorter tokens in the test data result in higher fertility, requiring more tokens to represent the same sequence compared to WordPiece and SentencePiece.
Conversely, SentencePiece's longer tokens in the test data lead to a lower fertility score, as fewer tokens are needed to represent the same sequence.
It also maintains more consistent token lengths between vocabulary and test data.
WordPiece generally falls between BPE and SentencePiece in terms of both token length and fertility, showing a clear trade-off between token length and the number of tokens required for encoding.

\subsection{Contextual Exponence}
\label{section:contextual-exponence}

\begin{figure}[htbp]
    \centering
    \subfloat[]{\includegraphics[width = .48\columnwidth]{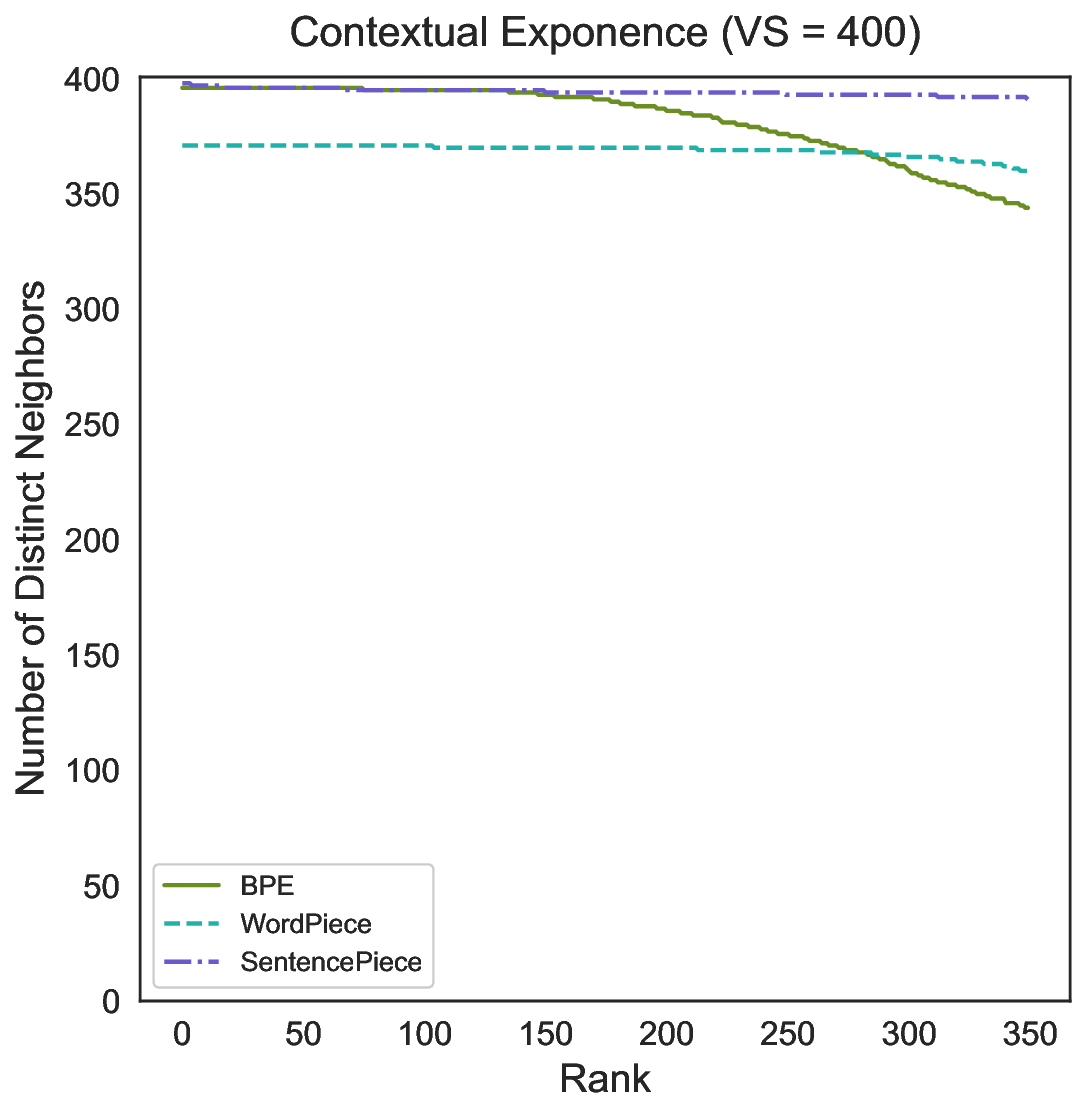}}\hfill
    \subfloat[]{\includegraphics[width = .48\columnwidth]{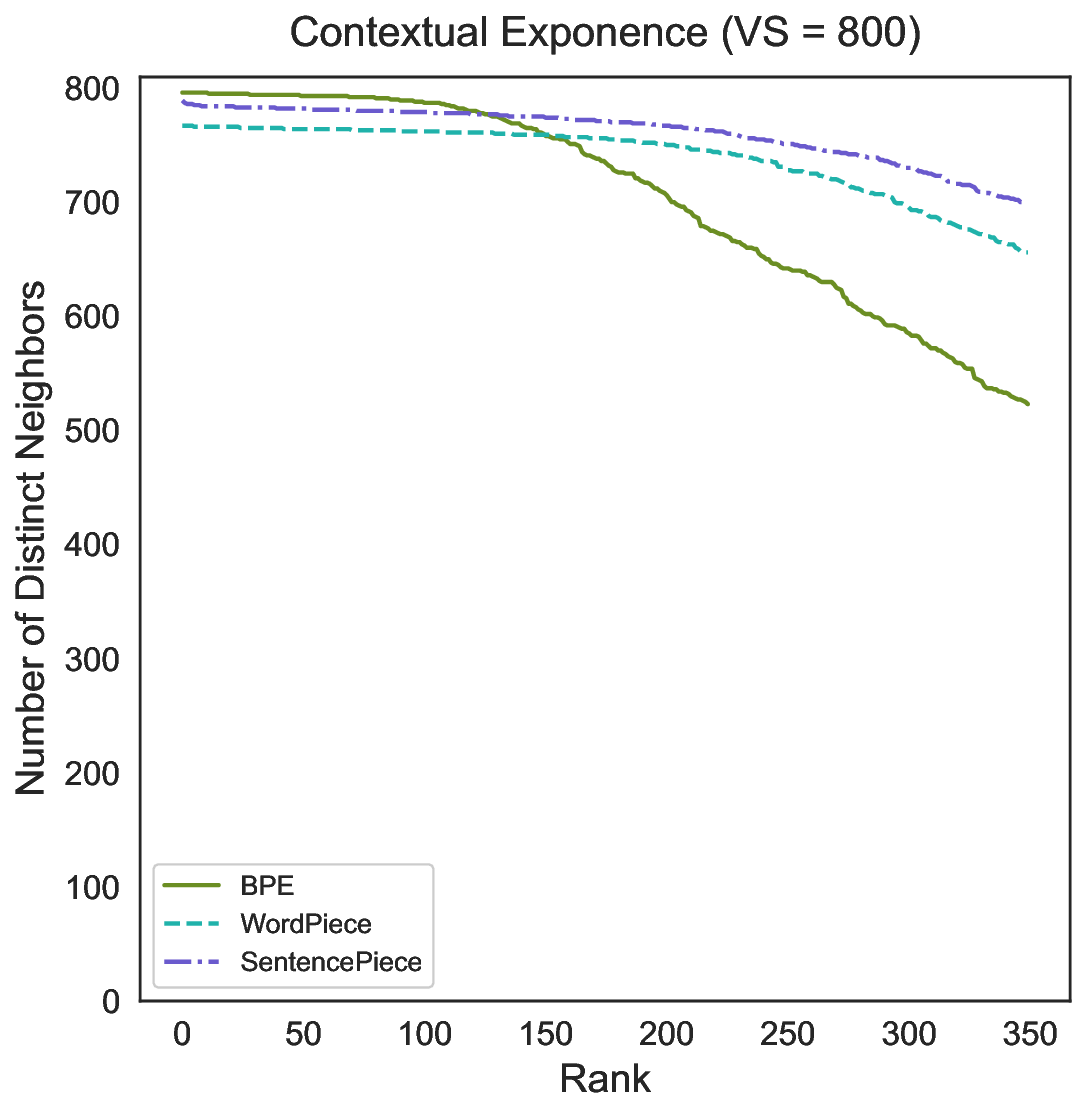}}
    
    \subfloat[]{\includegraphics[width = .48\columnwidth]{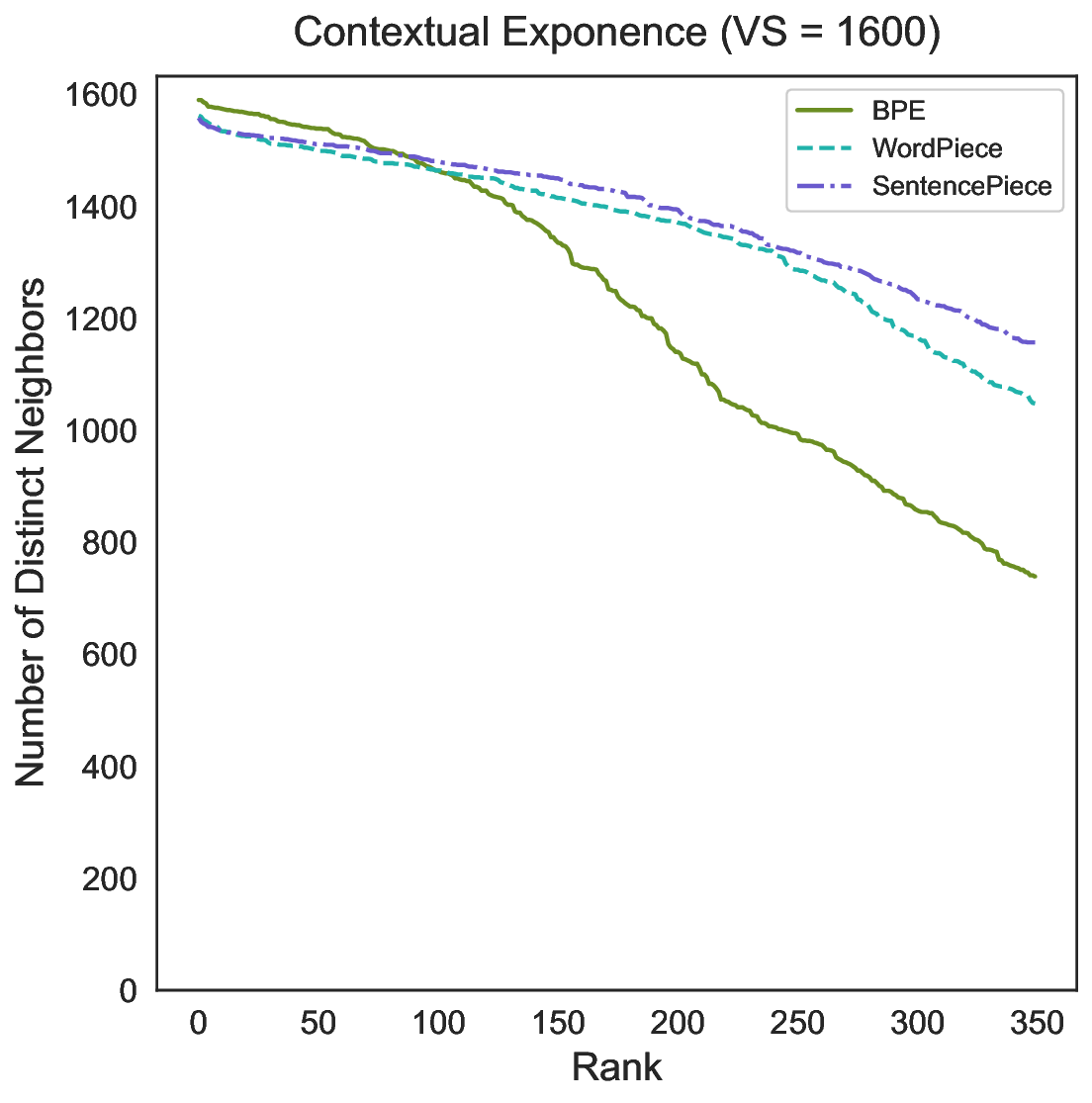}}\hfill
    \subfloat[]{\includegraphics[width = .48\columnwidth]{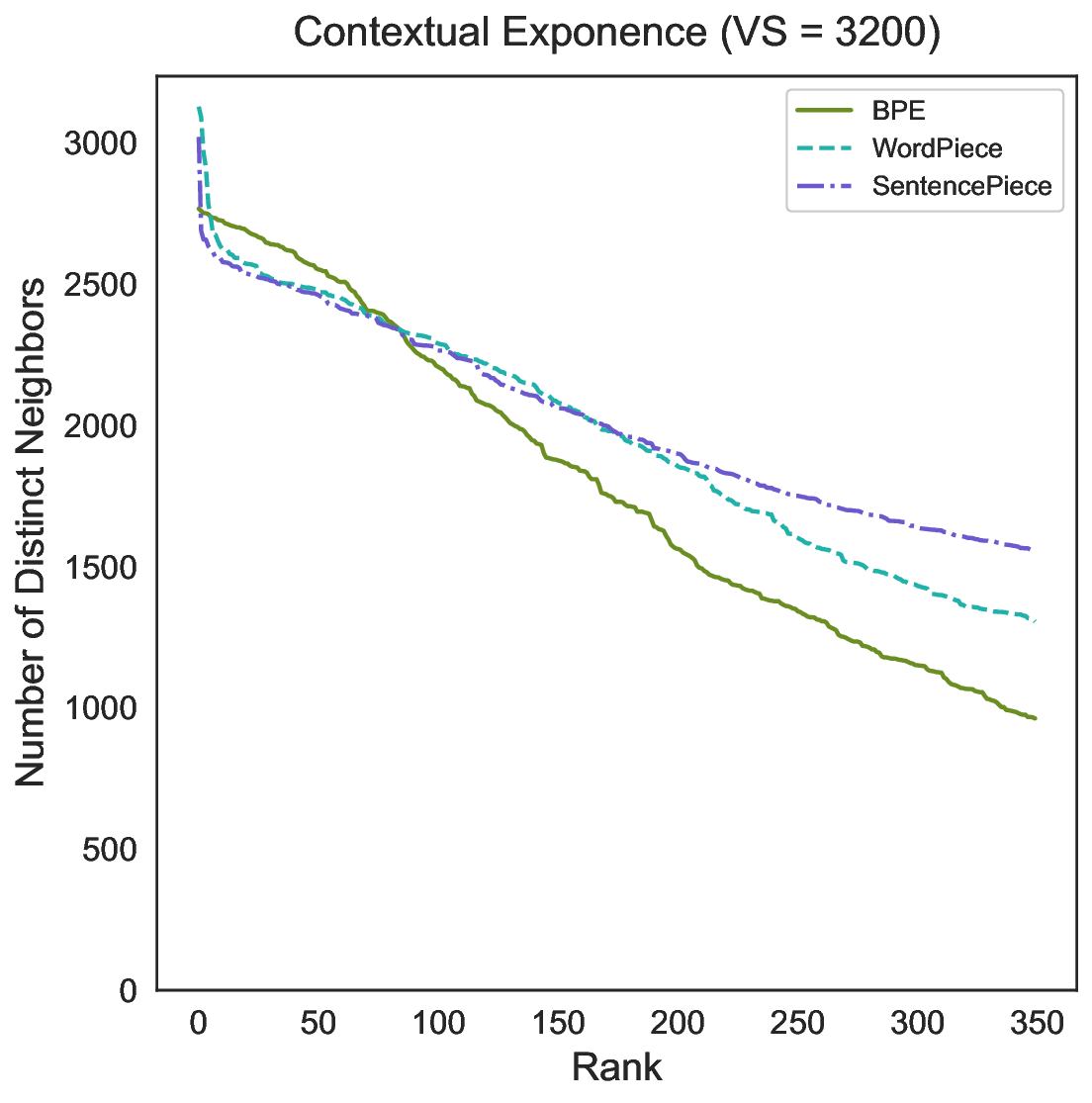}}
    
    \subfloat[]{\includegraphics[width = .48\columnwidth]{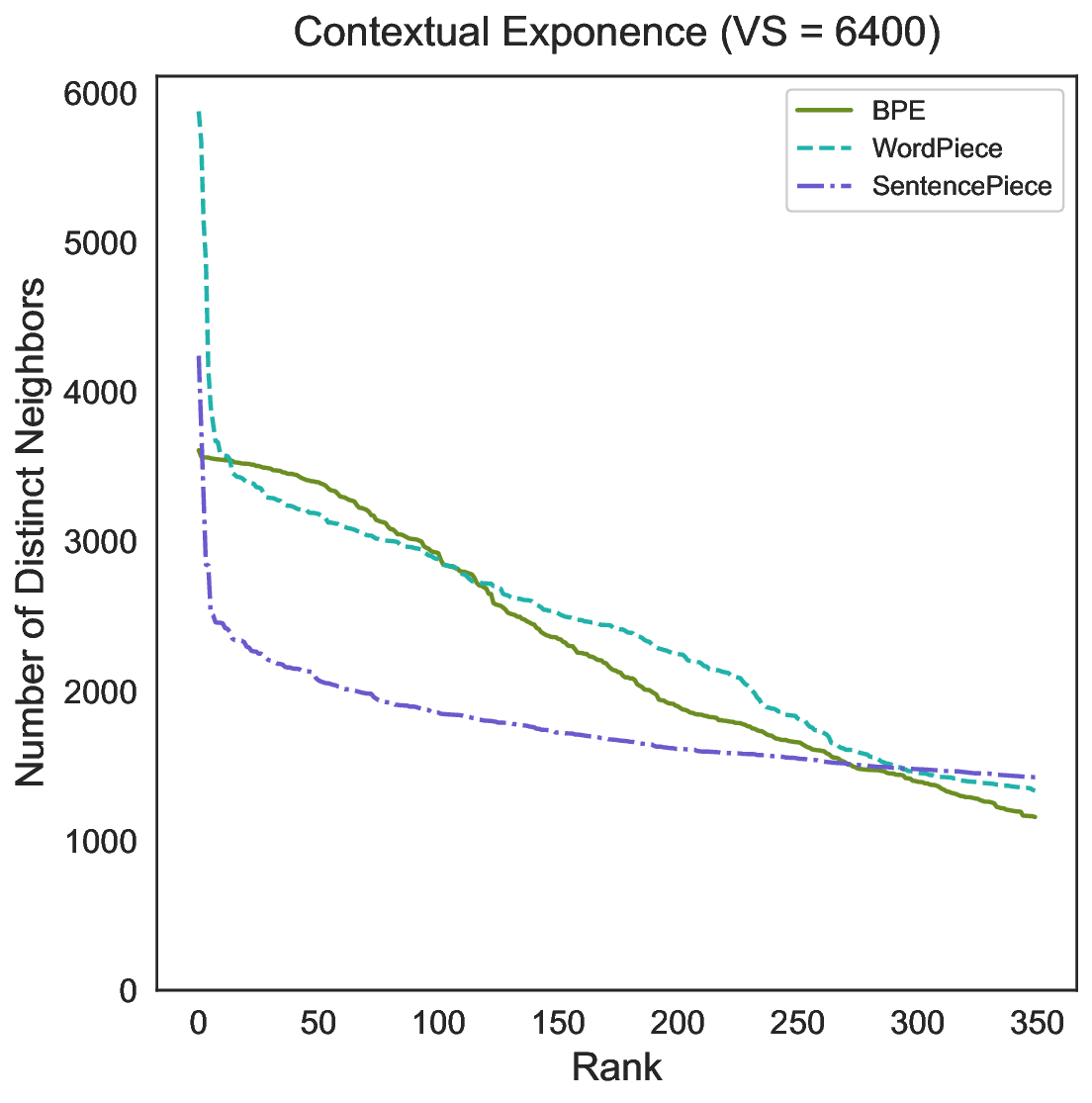}}
    \caption{Number of distinct neighbors each token encounters in a width-5 window, top 350 tokens. Plots are for BPE, WordPiece, and SentencePiece across different vocabulary sizes (VS).}
    \label{fig:ce_1}
\end{figure}

To evaluate how well tokenizers optimize tokens for contextual relevance, we visualize the number of unique neighbors each token encounters across the test data in Fig.~\ref{fig:ce_1}, ranked from highest to lowest \cite{yehezkel2023incorporating}.
Tokens designed to be context-independent  (appearing in nearly all contexts) dominate the top of the ranking.
For vocabulary size of 400, it is difficult to discern any clear patterns.
However, for vocabulary sizes of 800, 1600, and 3200, beyond the first hundred tokens, BPE's distinct neighbor counts fall below those of WordPiece and SentencePiece.
This pattern persists throughout the vocabulary, yielding a more contextually consistent token set for BPE.
At vocabulary size of 6400, SentencePiece assigns similar distinct neighbor counts to each token, flattening its plot. Meanwhile, WordPiece begins to behave similarly to BPE, with their distinct neighbor counts falling below those of SentencePiece after the first 300 tokens.

\subsection{Protein Domain Boundary Alignment}
\label{section:domains}

Protein domains are distinct functional and structural units within a protein sequence with well-defined boundaries. These regions often correspond to specific biological functions, and their integrity is crucial to the protein's overall structure and function. Protein domains can be considered analogous to linguistic phrases within a sentence: cohesive substructures that should not be split arbitrarily. Maintaining domain boundaries during tokenization is essential to preserving the biological meaning of the sequence and ensuring accurate downstream tasks such as structure prediction or functional annotation.

In this study, we evaluated how well tokenization methods respect the boundaries of protein domains. Using the PROSITE database~\cite{sigrist2012new}, we identified 4646 domains in 3377 of our test protein sequences. For each tokenized sequence, we examined whether the start of a domain coincided with the beginning of a token and the end of a domain aligned with the end of a token. We consider it a hit for the domain if both conditions are satisfied. Ideally, domain boundaries should not fall within the middle of any token.

\begin{figure}[htbp]
	\begin{center}
		\includegraphics[width=.85\columnwidth]{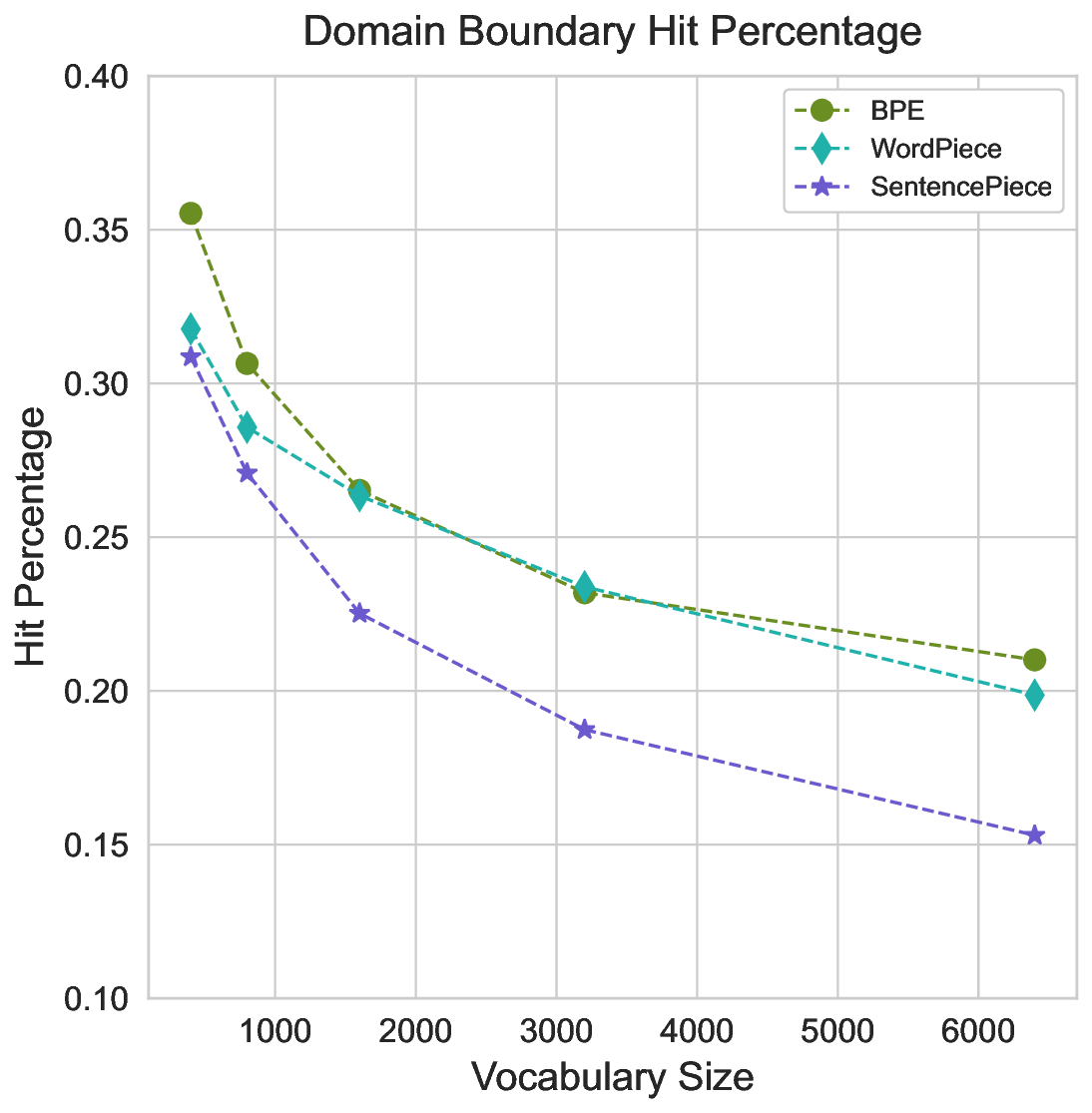}
		\caption{A domain is considered a hit if its start and end align with the beginning and end of a token, respectively. The plots show the hit percentages for BPE, WordPiece, and SentencePiece across different vocabulary sizes.}
		\label{fig:domain_boundary}
	\end{center}
\end{figure}

Our results in Fig.~\ref{fig:domain_boundary} show that the BPE tokenizer performs best in maintaining the protein domains' start and end boundaries. However, we also observed a decline in performance as the vocabulary size increased. This can be attributed to the fact that larger vocabularies result in longer tokens, making it harder for tokenization to respect domain boundaries accurately. Another critical observation is that BPE's relatively higher performance may not be solely due to its algorithm but rather to the shorter average token lengths it generates compared to other methods. As demonstrated earlier in Fig.~\ref{subfig:token_len_fertility_a2}, BPE tends to produce shorter tokens when applied to protein sequences, which naturally aligns better with domain boundaries.
This correlation between token length and domain boundary alignment holds for every tokenizer and vocabulary size.
While this result is expected, it raises a critical point. If tokenizers were truly capturing the underlying linguistic units of protein sequences, we would expect performance to remain stable with increasing vocabulary size. The drop in performance for larger vocabularies combined with the relatively low accuracy even for small vocabularies suggests that the tokenizers, which were originally developed for natural (human) languages, do not effectively capture the true linguistic subunits of protein sequences.

\subsection{Zipf's Law}
\label{section:zipfs-law}

\begin{figure}[htbp]
	\begin{center}
		\includegraphics[width=.85\columnwidth]{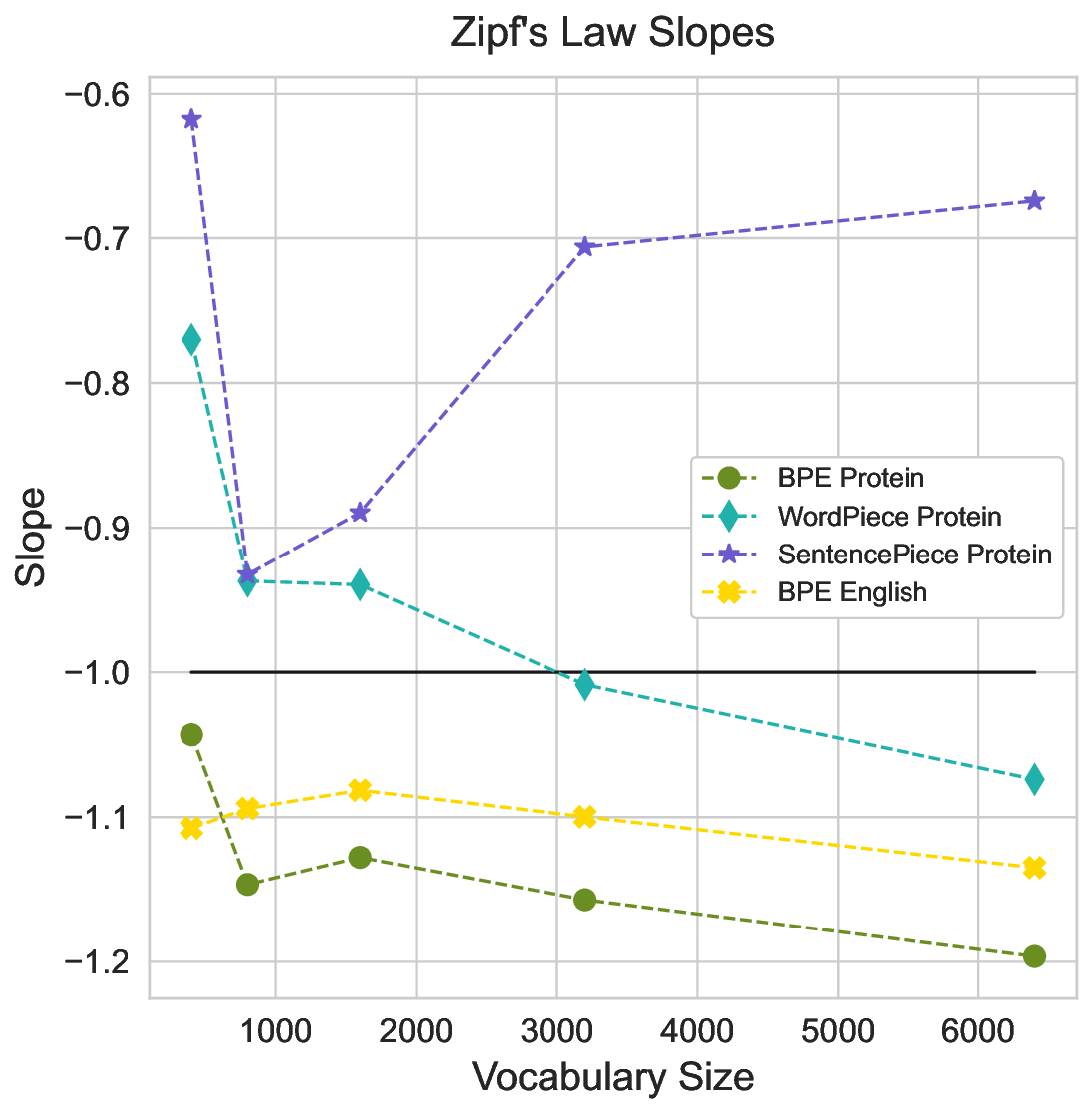}
		\caption{The slope values for Zipf's law plots of BPE (Protein and English), WordPiece (Protein), and SentencePiece (Protein) across different vocabulary sizes. -1 is the ideal slope value.}
		\label{fig:zipf_1}
	\end{center}
\end{figure}

Zipf's law is a statistical principle that describes the distribution of frequencies of elements in a dataset.
Zipf's law states that the frequency of a particular element is inversely proportional to its rank~\cite{zipf1949human}.
In simpler terms, it suggests that a few elements occur frequently, while most elements occur infrequently.
To observe Zipf's law we plot the frequency of each token as a function of its frequency rank in a log-log scale where the ideal line has a slope of \(-1\).
Fig.~\ref{fig:zipf_1} shows how the slopes of Zipf's law plots for different tokenization methods change as the vocabulary size varies.

For BPE applied to protein sequences, we find that the slope is consistently slightly steeper than the ideal Zipfian slope of \(-1\), averaging around \(-1.15\). This indicates a distribution skewed more toward frequent tokens.
WordPiece starts above \(-1\) for smaller vocabularies but dips closer to \(-1\) as the vocabulary grows, suggesting it aligns more closely with a Zipfian distribution at larger vocabulary sizes.
SentencePiece starts with an initial slope near \(-0.6\), indicating a flatter distribution with fewer high-frequency tokens. As the vocabulary size increases, it appears to approach \(-1\), but eventually diverges back to \(-0.6\).

When applied to English text, BPE aligns well with Zipf's law across all vocabulary sizes. 
This consistency suggests that BPE is well-suited for natural language data.
In contrast, tokenization for protein sequences is more variable.
WordPiece closely aligns with the ideal Zipfian distribution, followed by BPE, while SentencePiece shows greater deviation from the ideal.

\subsection{Brevity Law}
\label{section:brevity-law}

\begin{figure}[htbp]
    \centering
    \subfloat[]{\includegraphics[width = .48\columnwidth]{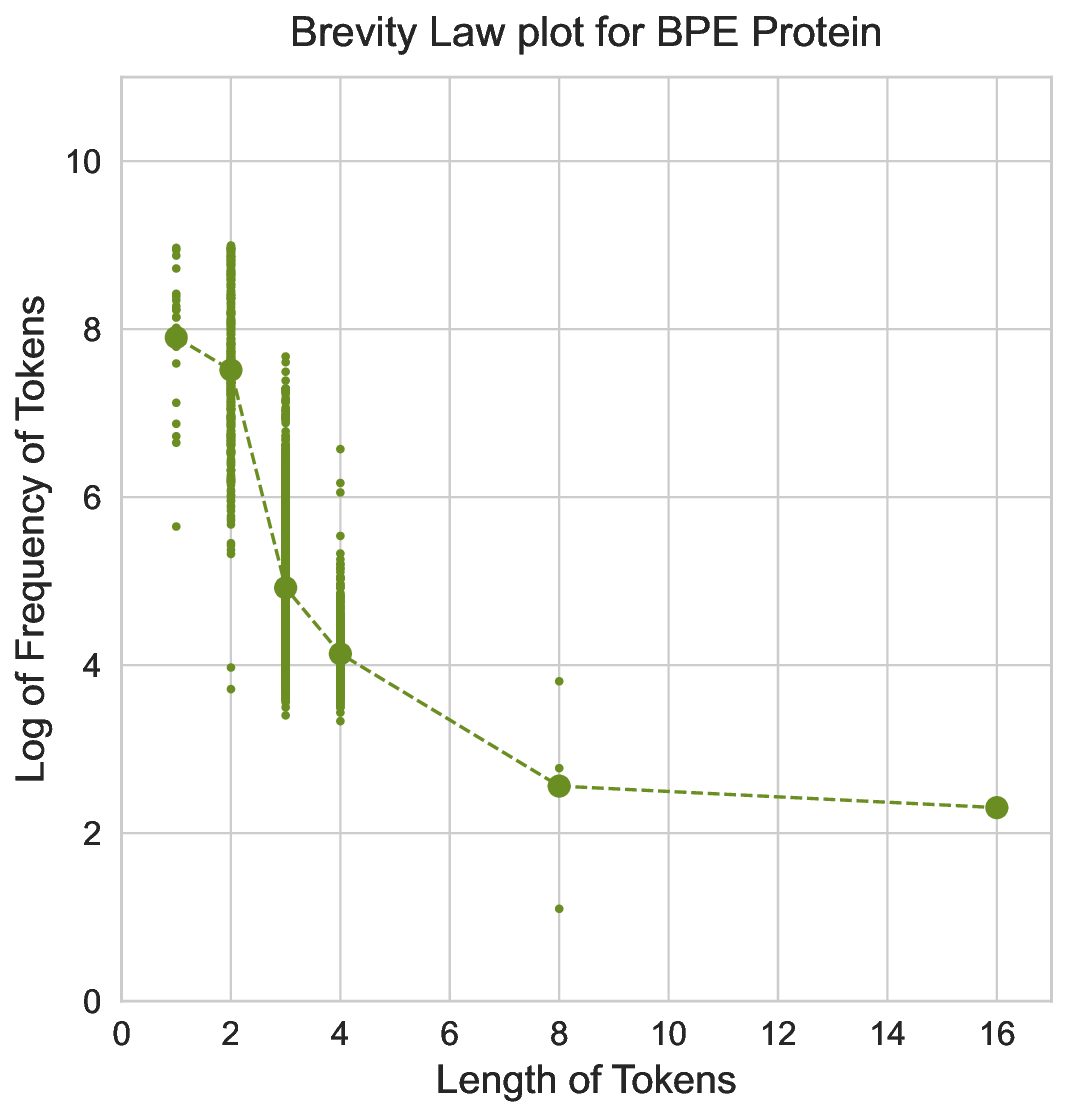}}\hfill
    \subfloat[]{\includegraphics[width = .48\columnwidth]{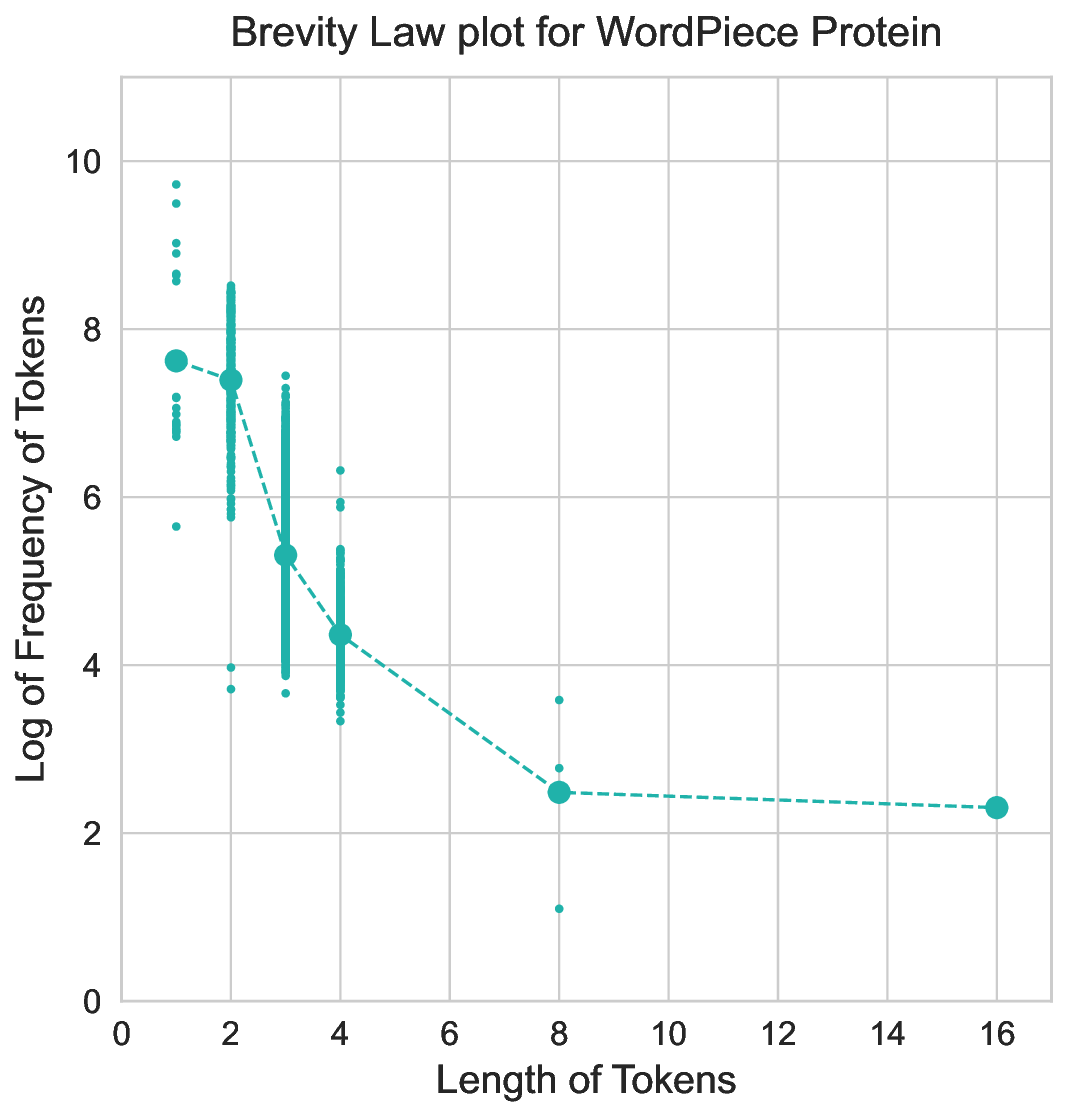}}
    
    \subfloat[]{\includegraphics[width = .48\columnwidth]{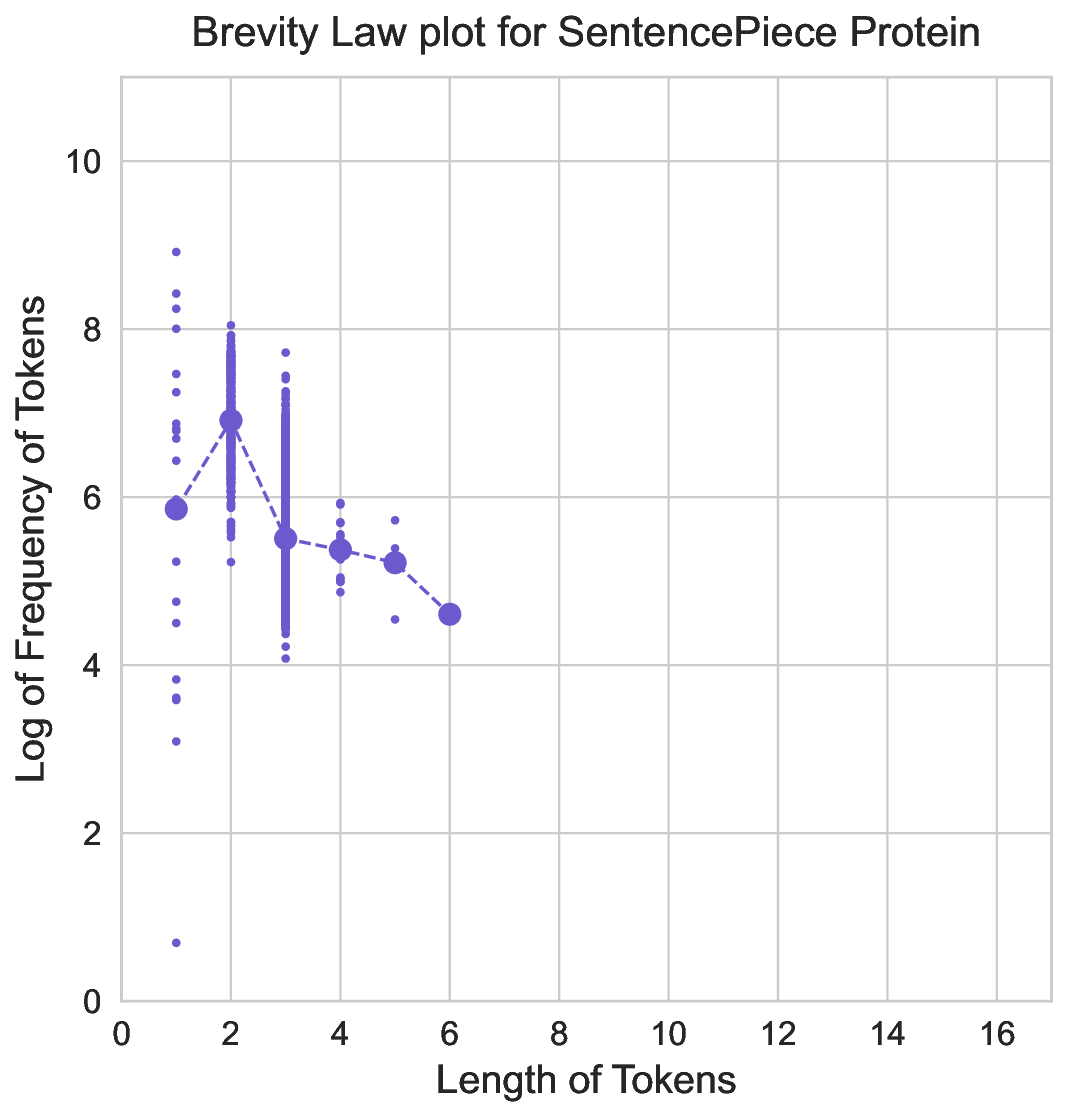}}\hfill
    \subfloat[]{\includegraphics[width = .48\columnwidth]{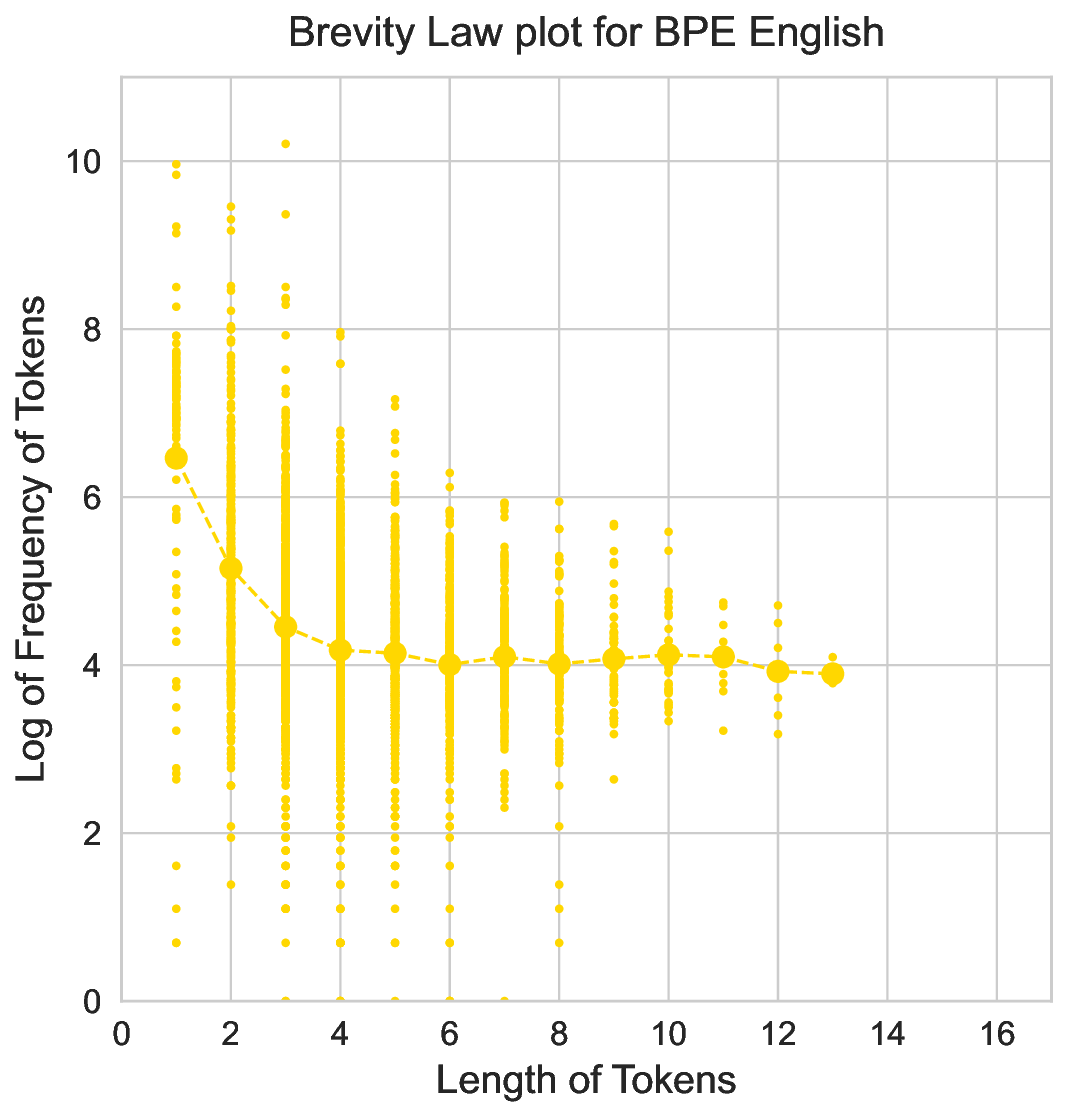}}
    \caption{Brevity law plots of BPE (Protein and English), WordPiece (Protein), and SentencePiece (Protein) for the vocabulary size of 3200.}
    \label{fig:brevity_1}
\end{figure}

\begin{figure}[htbp]
    \centering
    \subfloat[]{\includegraphics[width = .46\columnwidth]{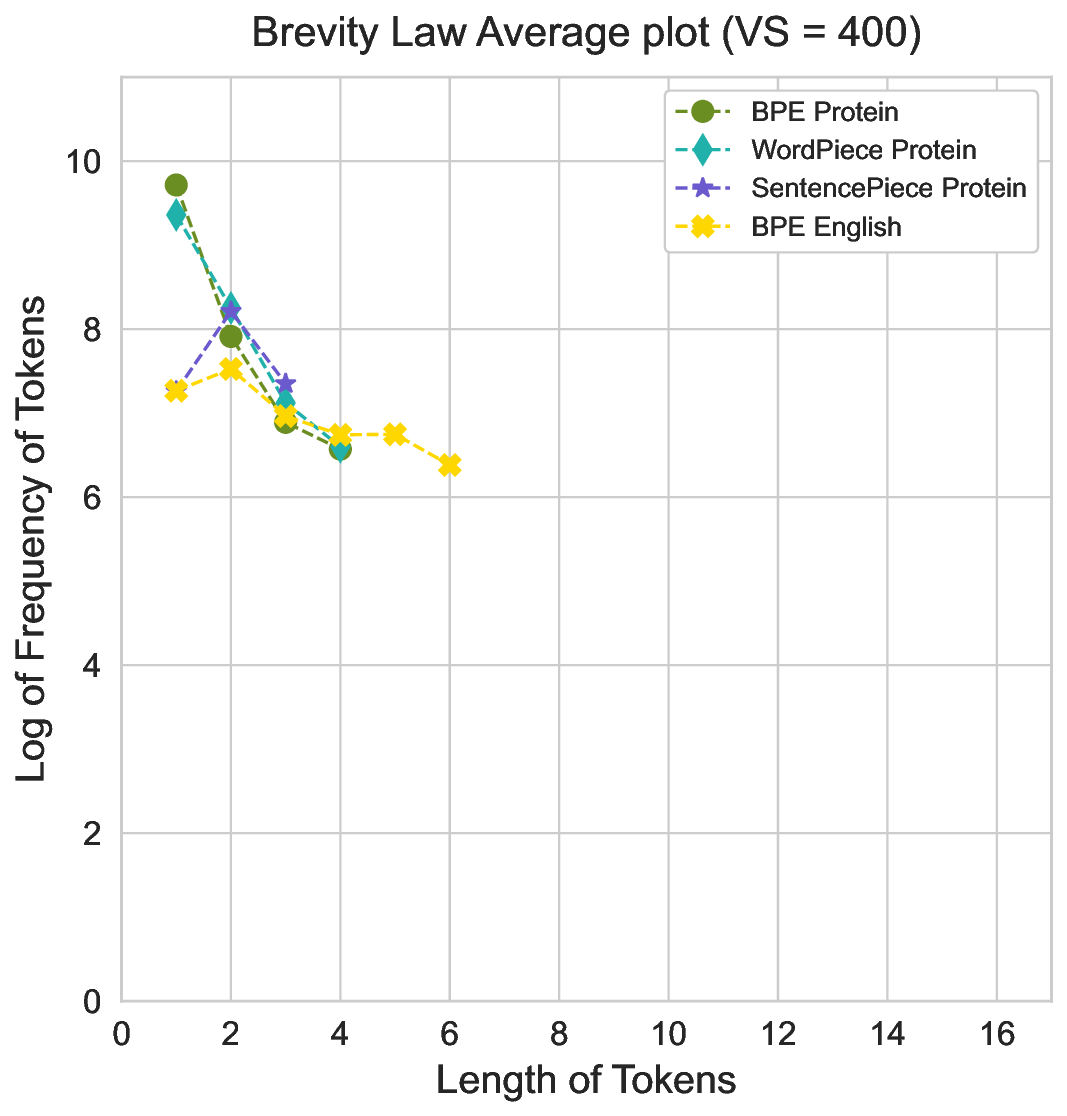}}\hfill
    \subfloat[]{\includegraphics[width = .46\columnwidth]{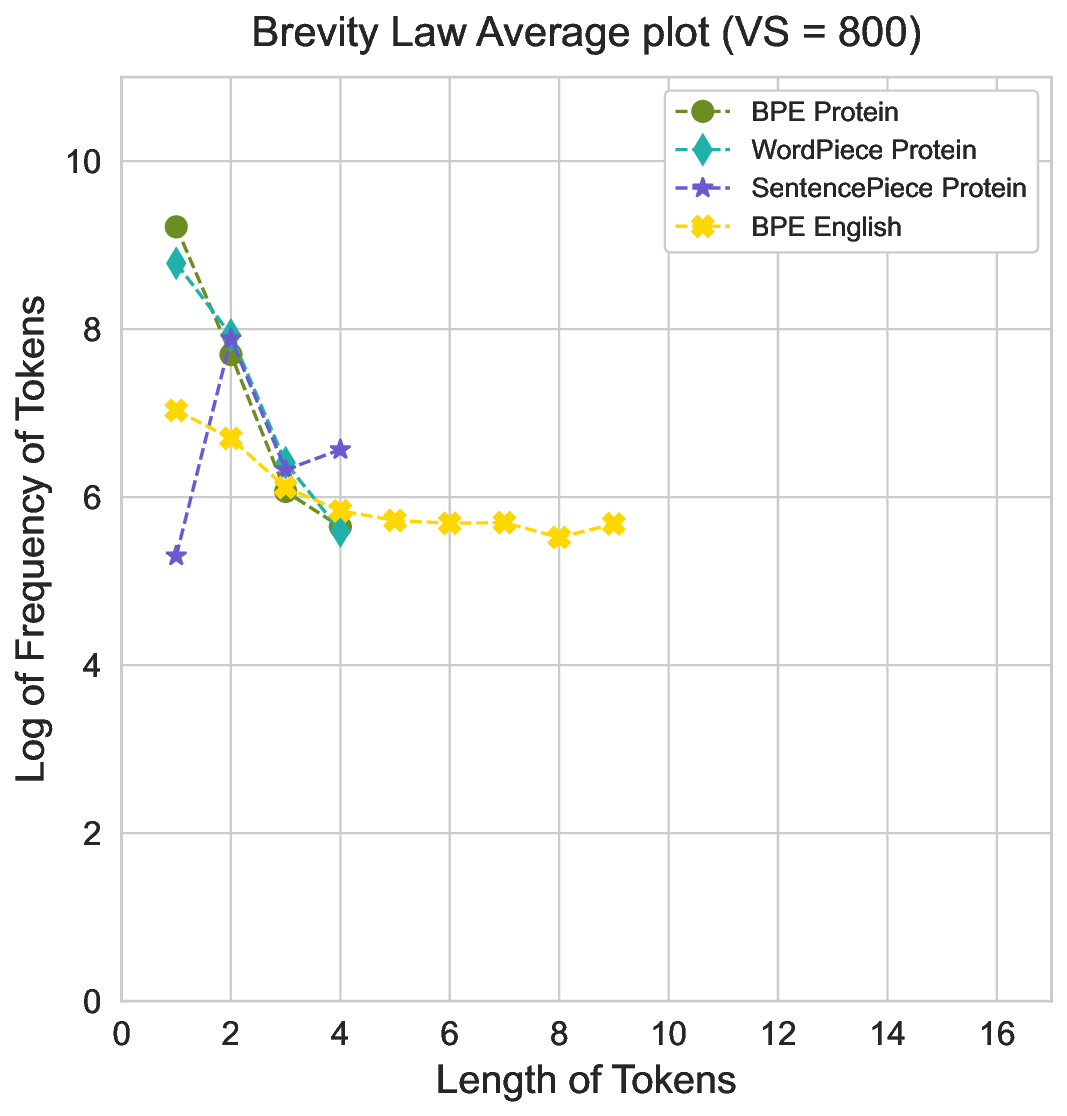}}
    
    \subfloat[]{\includegraphics[width = .46\columnwidth]{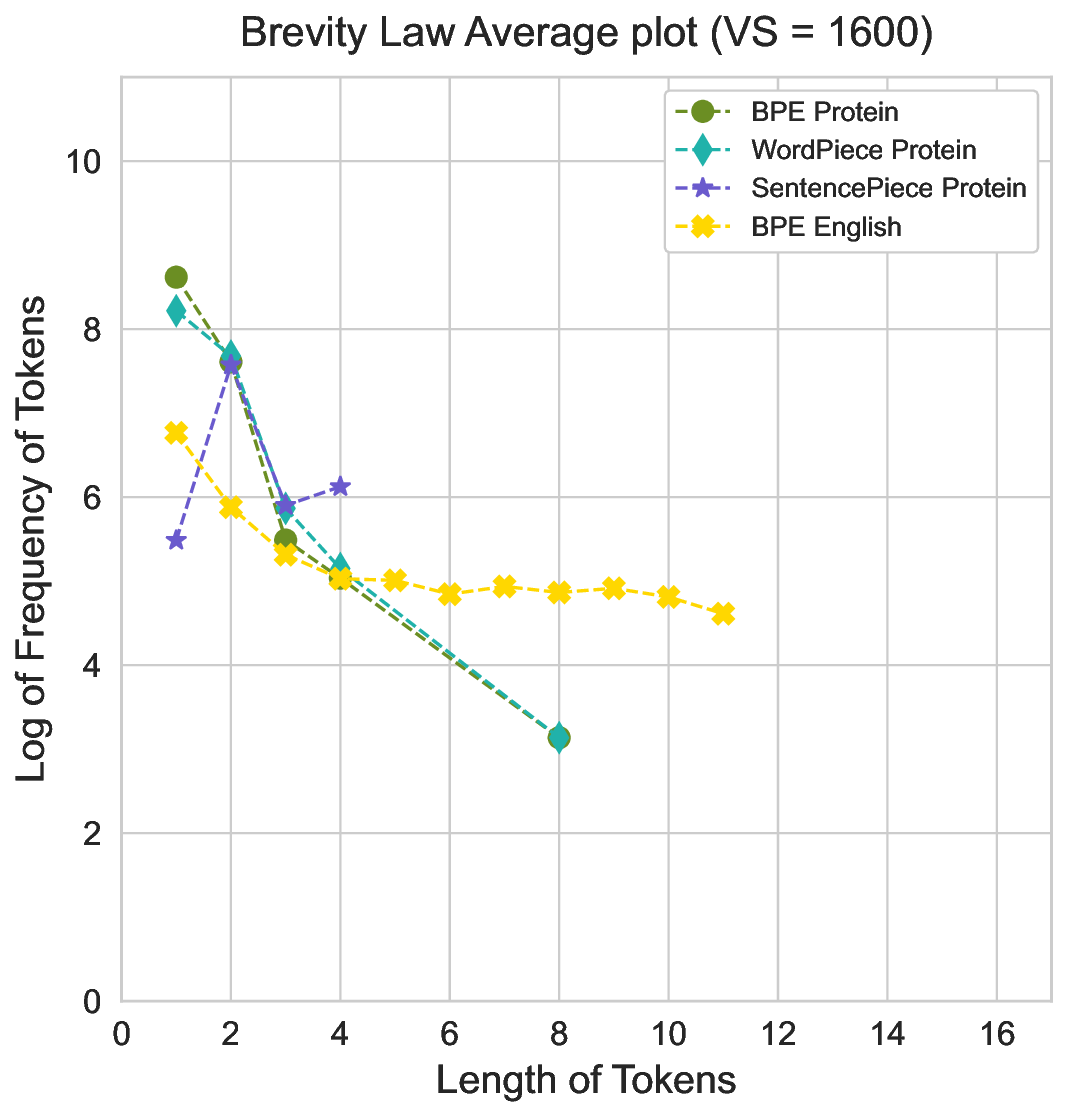}}\hfill
    \subfloat[]{\includegraphics[width = .46\columnwidth]{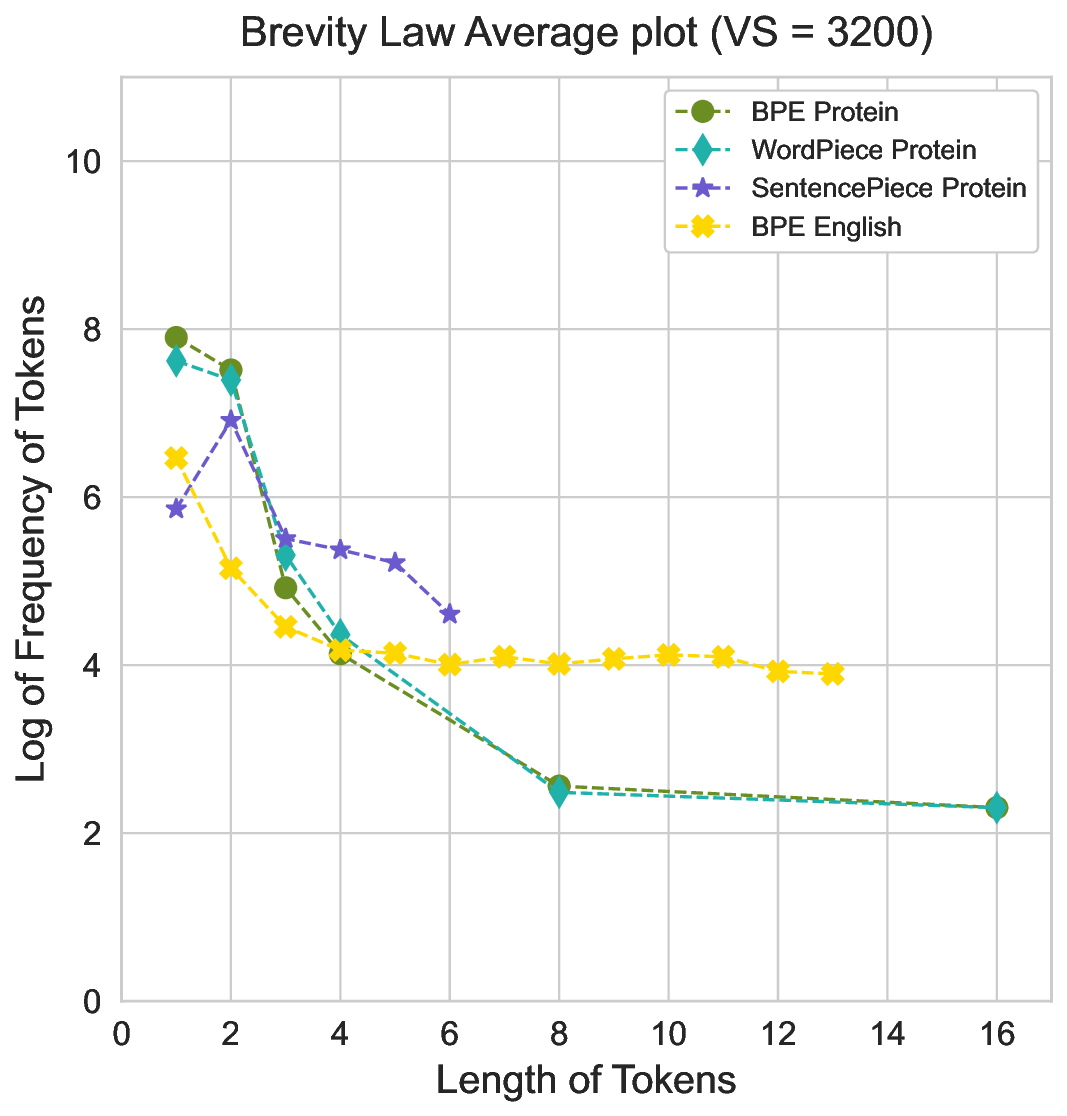}}
    
    \subfloat[]{\includegraphics[width = .46\columnwidth]{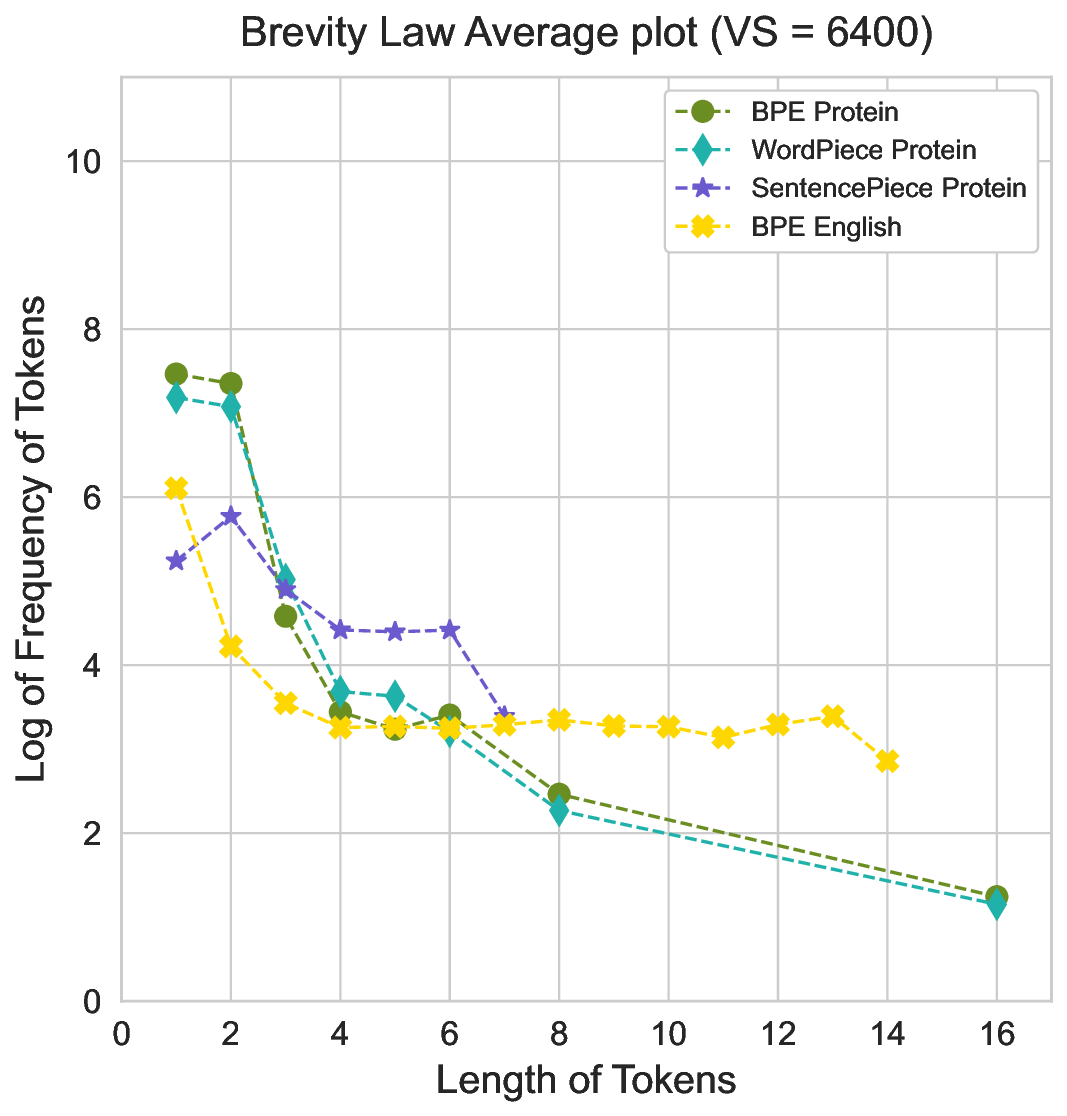}}
    \caption{Brevity law average plots of BPE (Protein and English), WordPiece (Protein), and SentencePiece (Protein) across different vocabulary sizes (VS).}
    \label{fig:brevity_2}
\end{figure}

Brevity law suggests that frequently used tokens tend to be shorter~\cite{zipf1949human, torre2019physical}, and we see that this pattern holds for different tokenizers, although each has its own approach to handling token lengths.
We plotted the log of the frequency of the tokens against the length of the tokens to observe the Brevity law. While Fig.~\ref{fig:brevity_1} demonstrates Brevity law plots for different tokenization methods for the vocabulary size of 3200, for the plots in Fig.~\ref{fig:brevity_2} we took the average of token frequency for each token length to show all tokenizers for each vocabulary size.

For BPE and WordPiece applied to protein sequences, shorter tokens appear far more frequently, producing a steep drop in frequency as token length increases.
SentencePiece, on the other hand, displays more variation in shorter token frequencies, suggesting it has a slightly different segmentation strategy.

In English text, BPE shows a steady, predictable drop in frequency across token lengths, a reflection of its suitability for natural language. This consistency doesn’t quite translate to protein data, where the token length distributions are generally less stable. With larger vocabularies, however, we notice that BPE and WordPiece achieve a more refined alignment with Brevity law, producing fewer longer tokens, whereas SentencePiece retains more variability across lengths.
Overall, Brevity law is evident in all tokenization methods, with BPE and WordPiece seeming to adhere more closely to it and larger vocabularies leading to more consistent tokenization patterns.

\subsection{Heap's Law}
\label{section:heaps-law}

Heap's law is an empirical formula stating that as dataset size increases, vocabulary size also increases, but at a decreasing rate~\cite{10.5555/539986}.
Heap's law is often used to estimate the vocabulary size needed for information retrieval systems or assess the vocabulary richness in a given text.
Its formula is expressed as:
\begin{equation}
V(n) = K \cdot n^\beta
\end{equation}
where $V(n)$ is the estimated vocabulary size when the document or collection contains $n$ words, $K$ is a constant, typically in the range of 10 to 100, and $\beta$ is an exponent, typically in the range of 0.4 to 0.6.

\begin{figure}[htbp]
    \centering
    \subfloat[]{\includegraphics[width = .48\columnwidth]{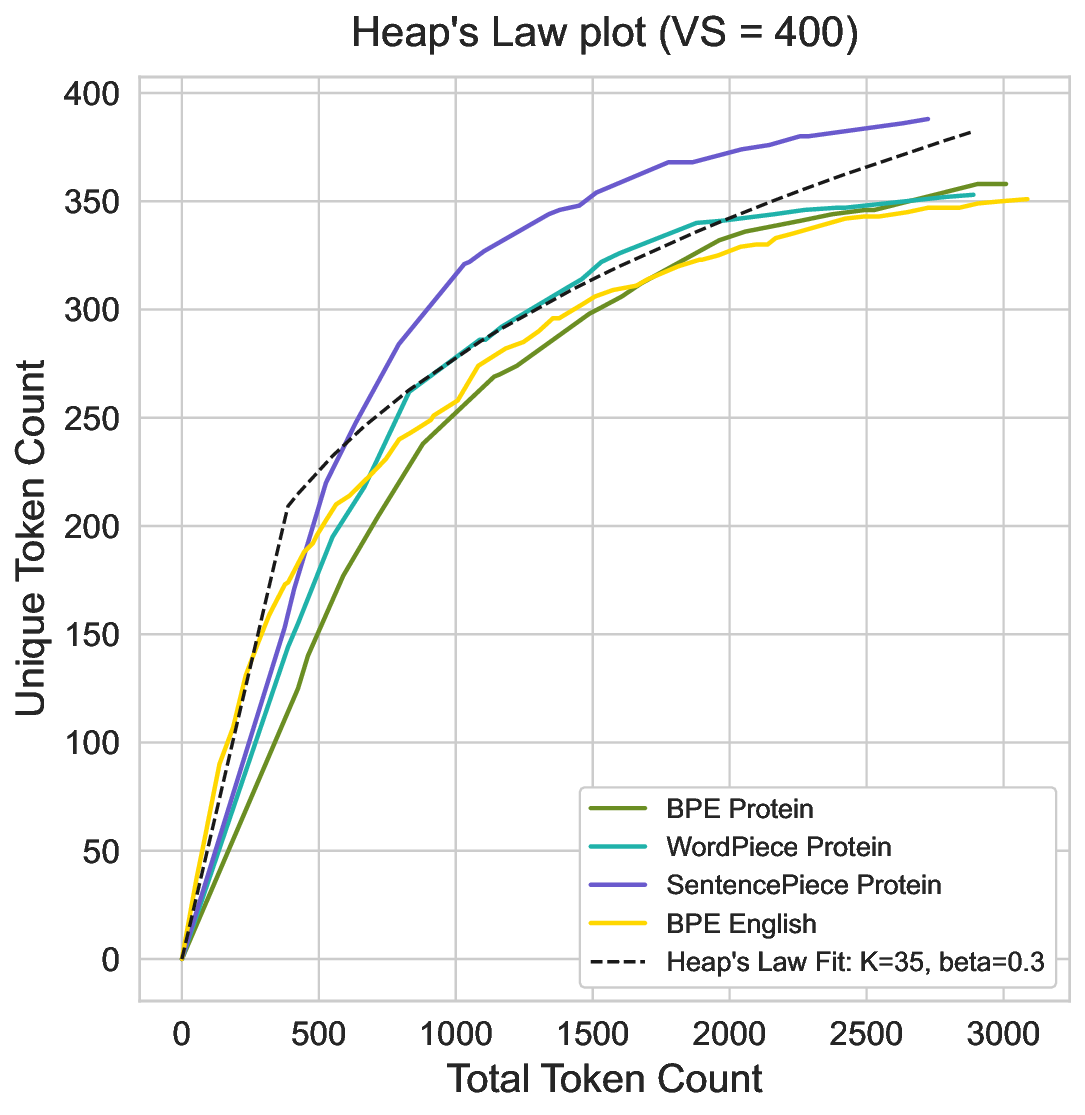}}\hfill
    \subfloat[]{\includegraphics[width = .48\columnwidth]{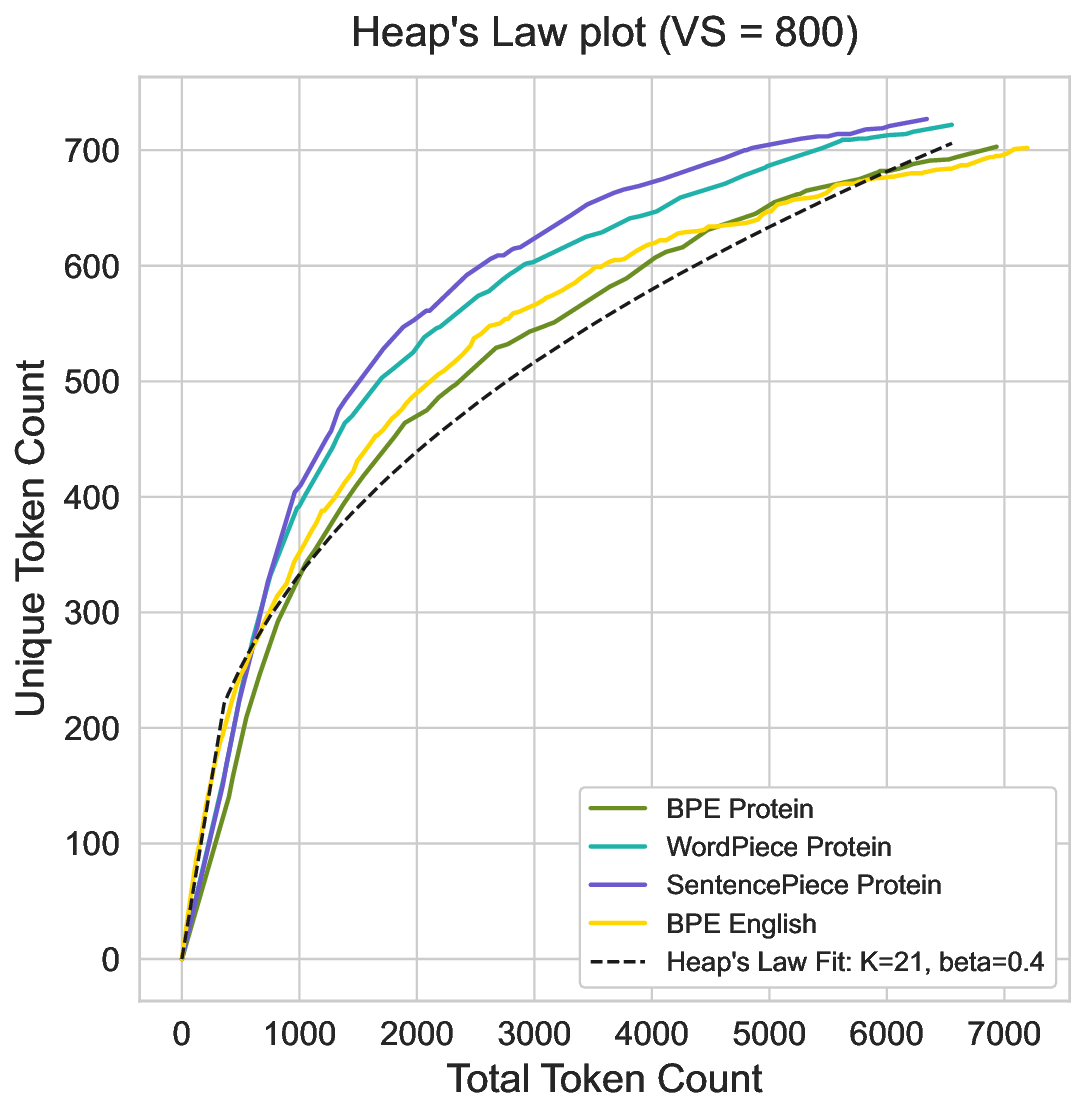}}
    
    \subfloat[]{\includegraphics[width = .48\columnwidth]{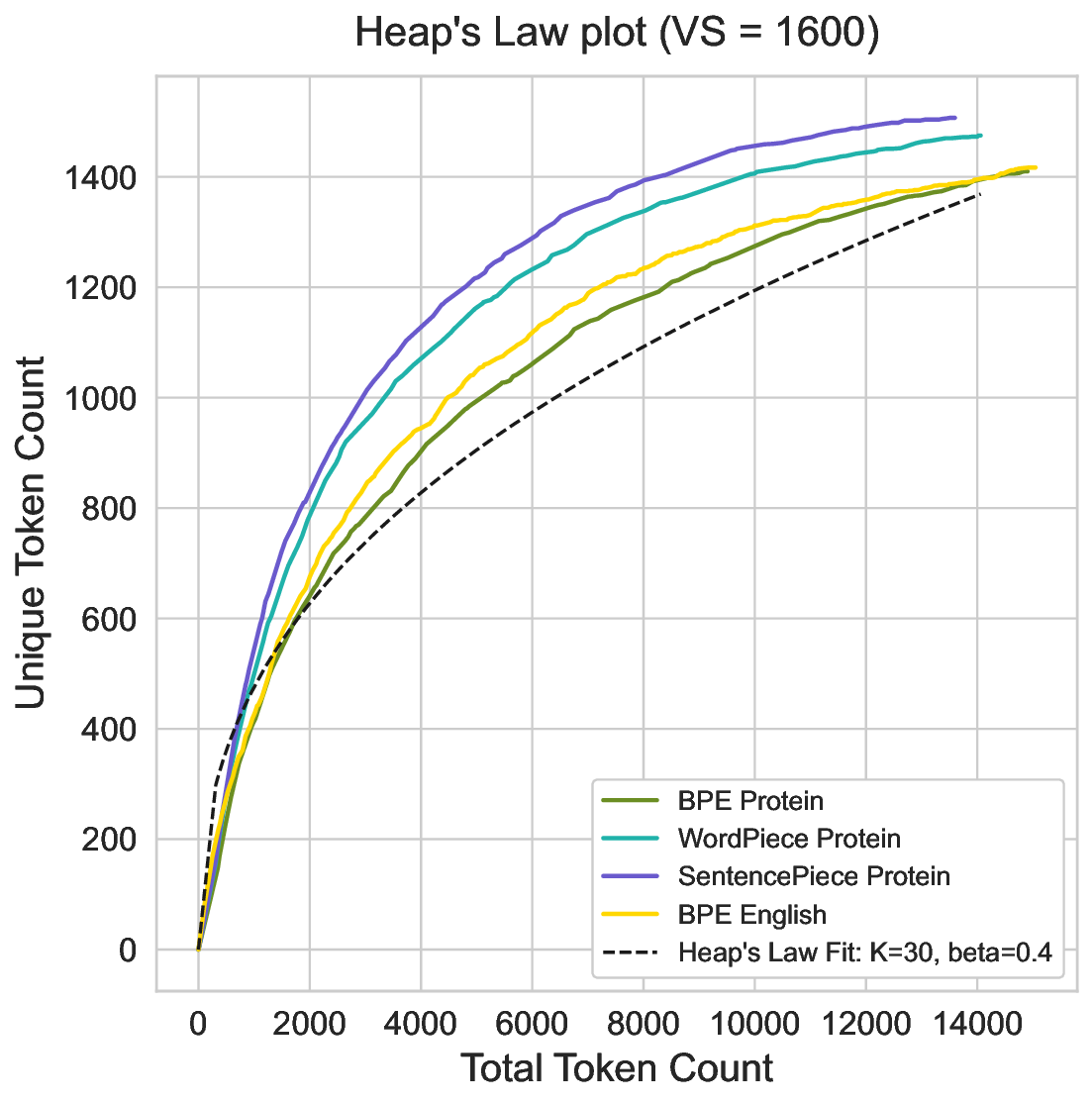}}\hfill
    \subfloat[]{\includegraphics[width = .48\columnwidth]{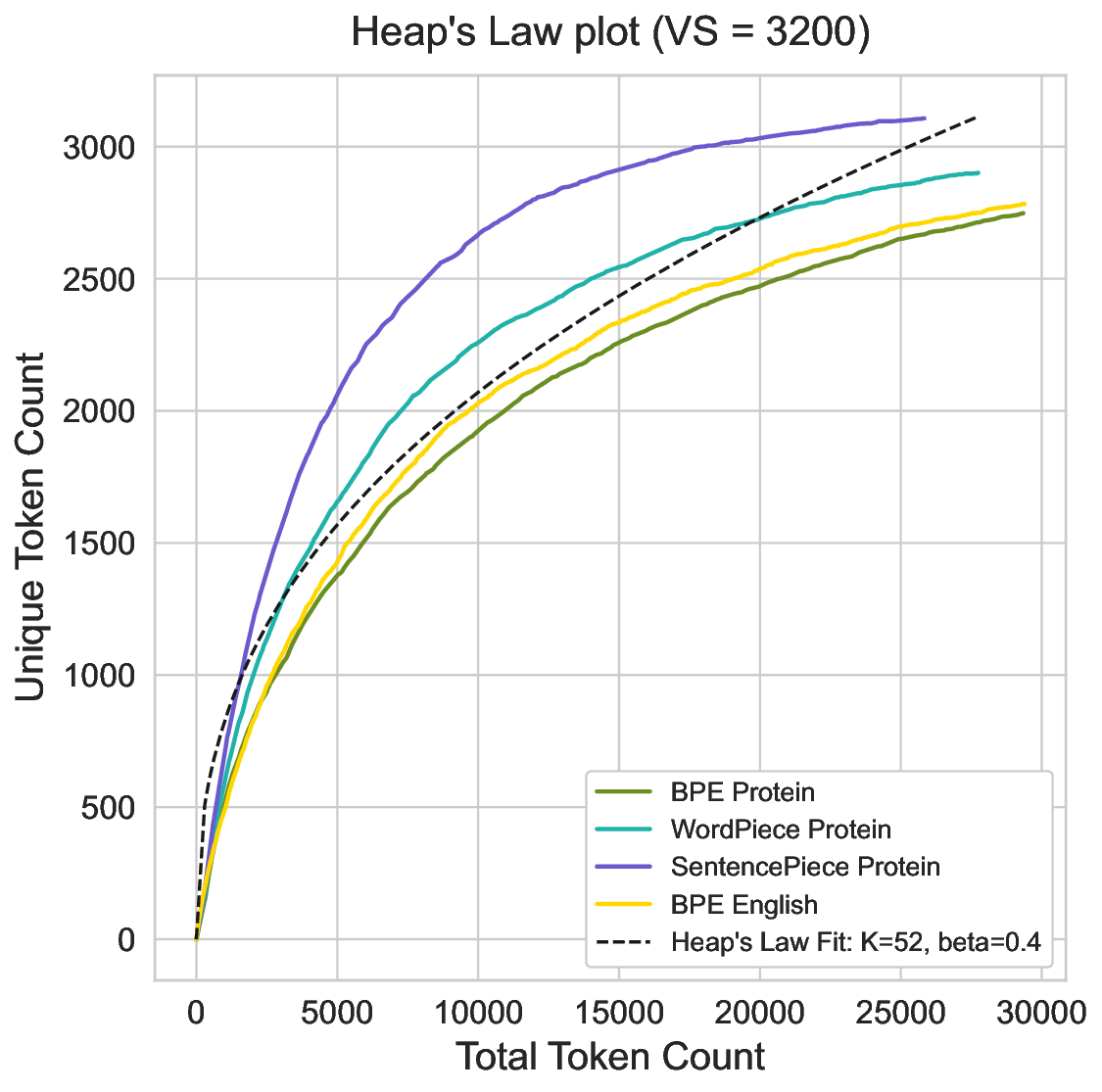}}
    
    \subfloat[]{\includegraphics[width = .48\columnwidth]{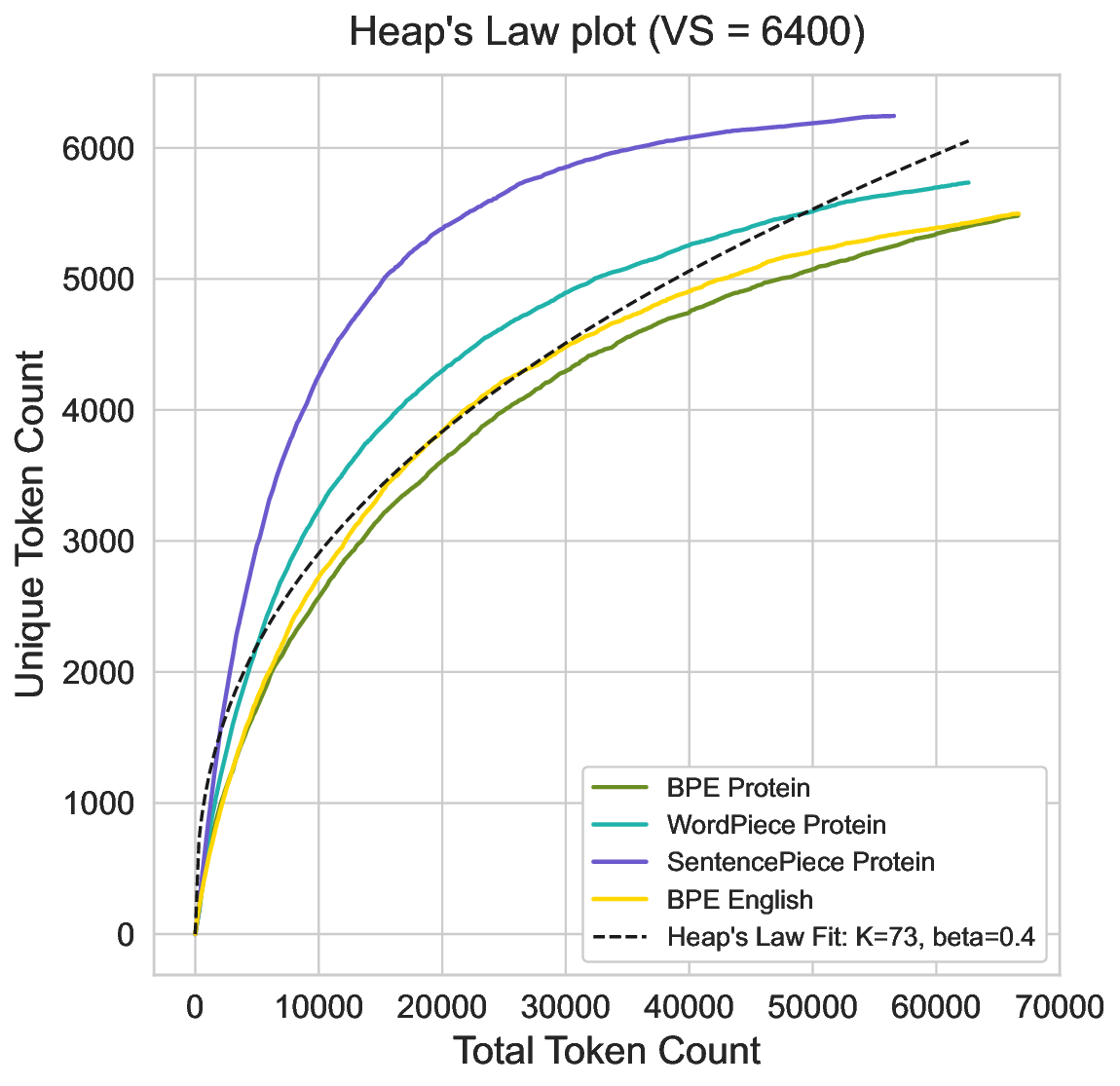}}
    \caption{Heap's law plots of BPE (Protein and English), WordPiece (Protein), and SentencePiece (Protein) across different vocabulary sizes (VS).}
    \label{fig:heap_1}
\end{figure}

The plots in Fig.~\ref{fig:heap_1} representing Heaps' law across different tokenizers and vocabulary sizes exhibit a typical behavior where the number of unique tokens increases with the total token count but at a diminishing rate.
Although SentencePiece appears to reach saturation slightly faster than the others, we can state that all tokenizers follow Heaps's law closely.

\subsection{Menzerath's Law}
\label{section:Menzeraths-law}

Menzerath's law, also known as Menzerath–Altmann law, states that larger linguistic constructs tend to have shorter constituents~\cite{altmann1980prolegomena}. In the context of sequence tokenization, this implies that as the sequence length increases, the average token length should decrease.
However, our findings show a more complex behavior for protein tokenizers, with none of the tokenizers fully following this expected pattern.

\begin{figure}[htbp]
    \centering
    \subfloat[]{\includegraphics[width = .48\columnwidth]{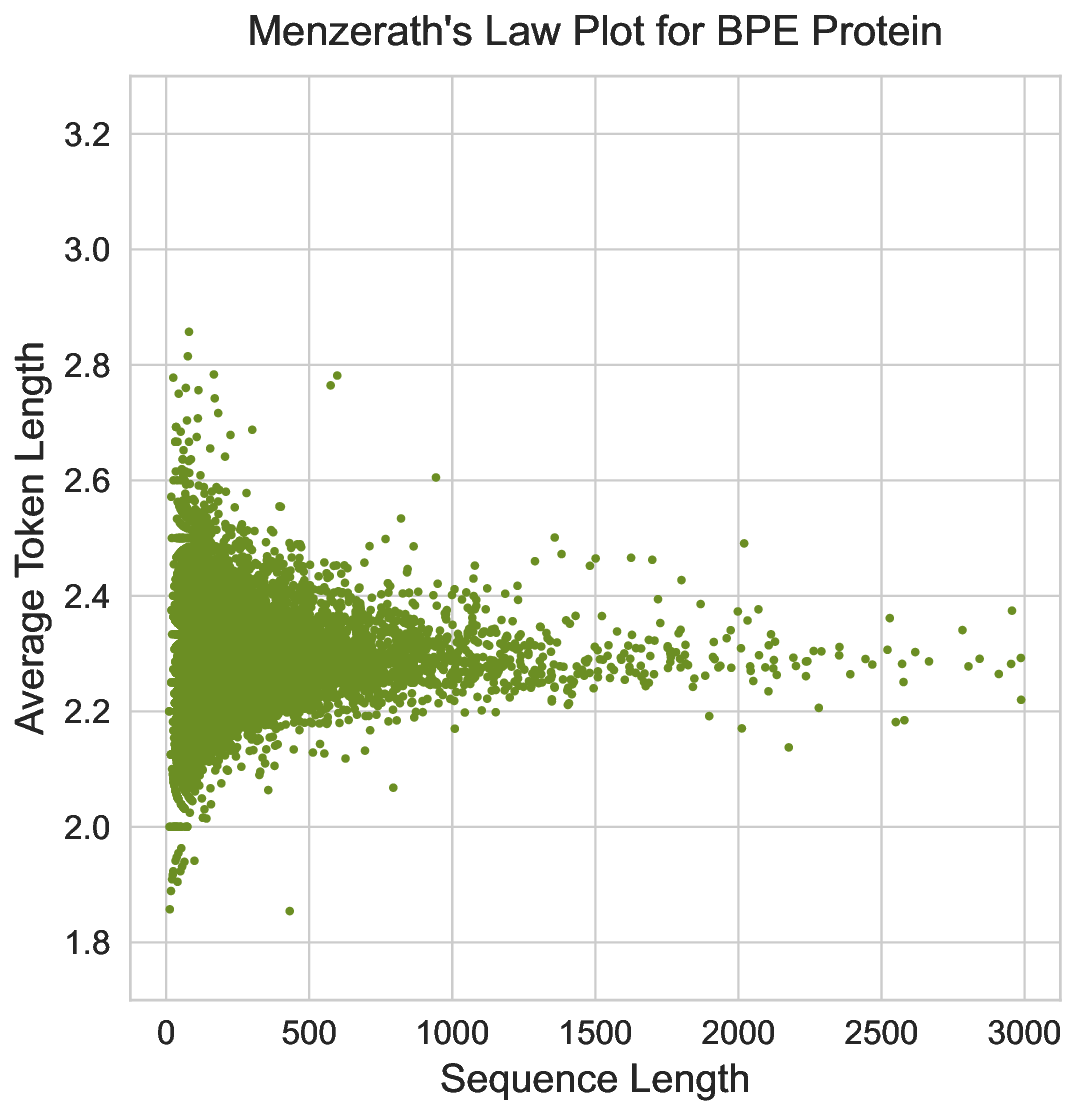}}\hfill
    \subfloat[]{\includegraphics[width = .48\columnwidth]{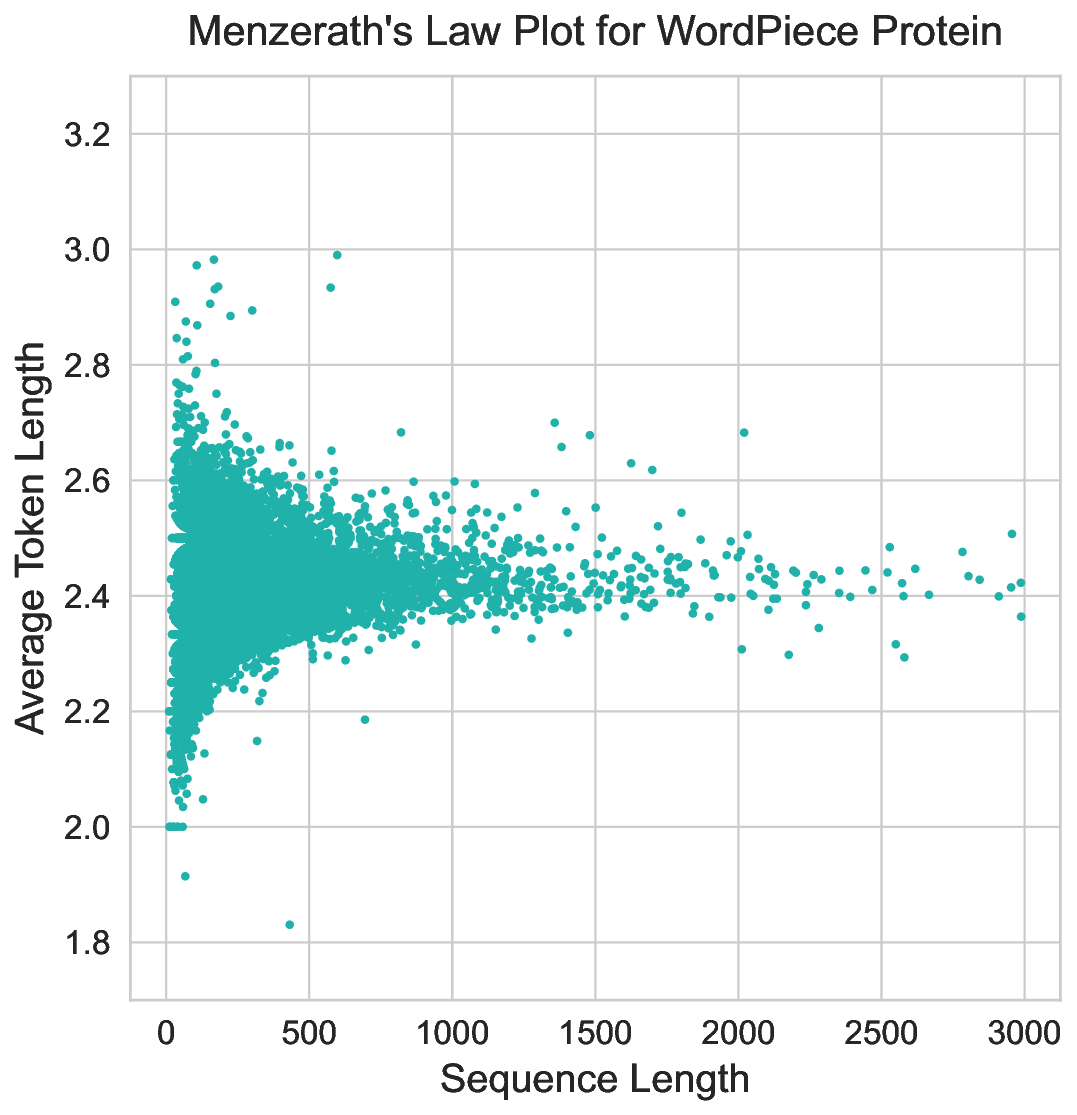}}
    
    \subfloat[]{\includegraphics[width = .48\columnwidth]{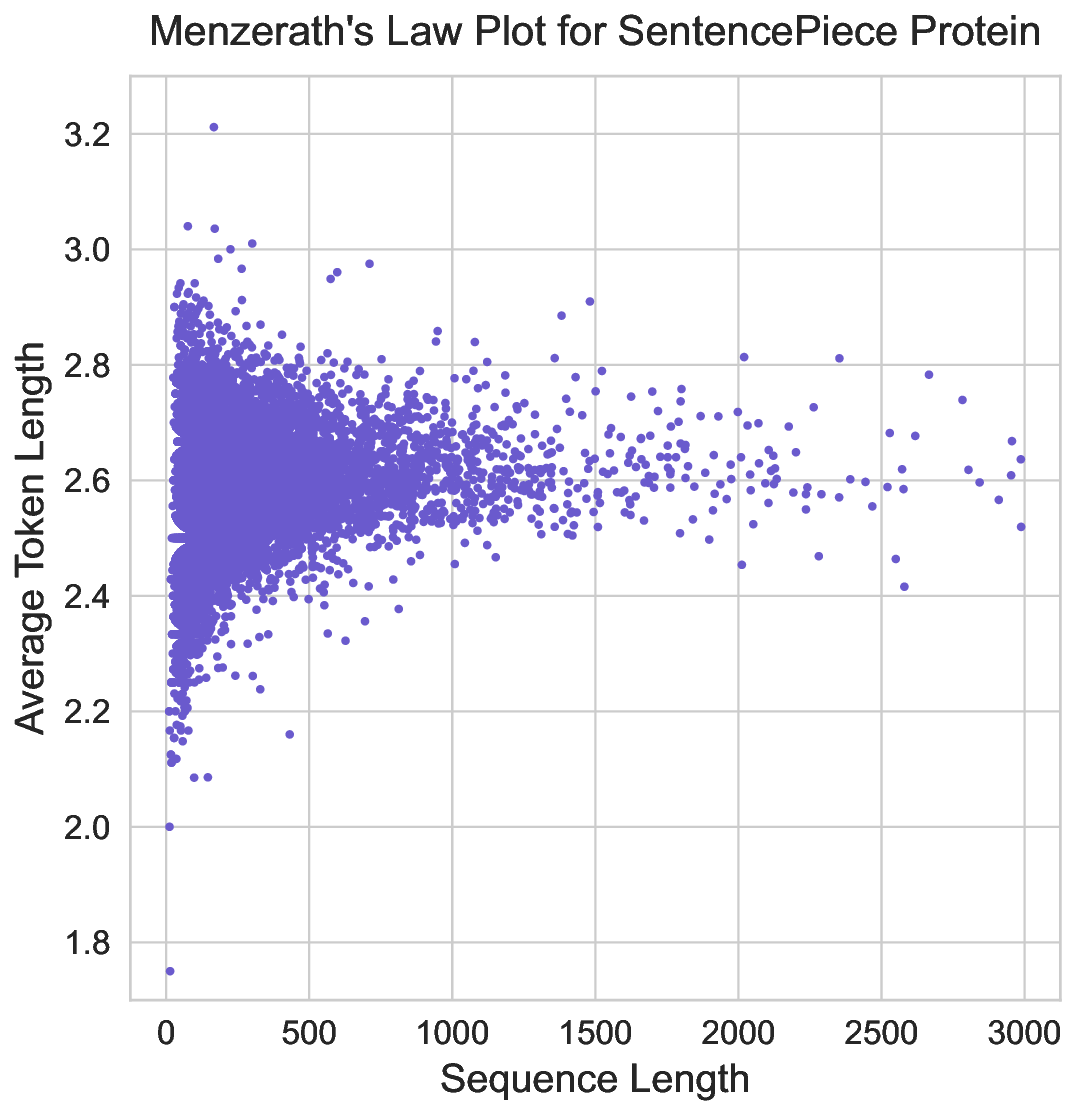}}\hfill
    \subfloat[]{\includegraphics[width = .48\columnwidth]{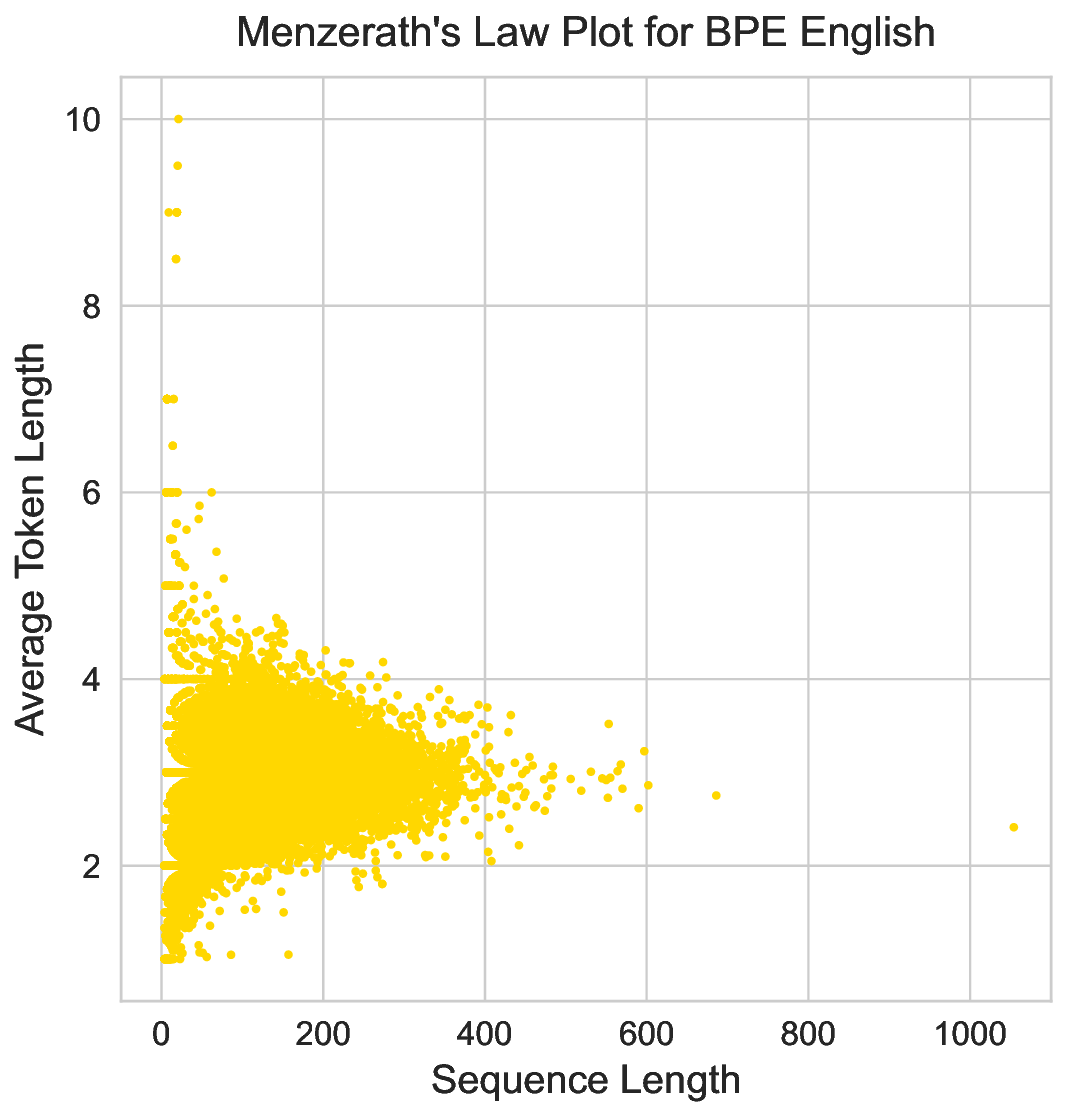}}
    \caption{Menzerath's law plots of BPE (Protein and English), WordPiece (Protein), and SentencePiece (Protein) for vocabulary size of 3200.}
    \label{fig:menzerath_1}
\end{figure}

Fig.~\ref{fig:menzerath_1} reveals that while average token length stays roughly constant as sequence length increases, there are simultaneous trends of both decreasing and increasing token lengths across longer sequences.
This mixed behavior suggests that, rather than fully adhering to Menzerath’s law, these tokenizers generate segments that vary in length.
When comparing these results with BPE applied to English text, we see two notable differences.
First, average token lengths decreases more consistently as sequences grow longer.
Second, there are fewer outliers compared to protein tokenizers, as the average token length for longer sequences in protein tokenizers tends to be more dispersed toward greater lengths compared to English text.

\section{Discussion and Conclusion}
\label{chapter:future-work}

In this study, we investigated whether NLP-based tokenizers, BPE, WordPiece and SentencePiece, can effectively capture the patterns and underlying structure of protein sequences. Through comprehensive analysis across various vocabulary sizes, we assessed their capabilities in representing the language of proteins and their compliance with established linguistic laws.

Our findings revealed distinct behavioral patterns among the tokenizers, with vocabulary size playing a crucial role in their performance. While smaller vocabularies produced relatively similar token sets across all methods, larger vocabularies led to significant divergence, particularly in SentencePiece's case. This observation suggests that both vocabulary size and tokenizer selection substantially impact protein sequence representation.

Each tokenizer demonstrated unique characteristics in handling protein sequences. BPE generated the longest vocabulary tokens but the shortest test tokens, resulting in high fertility and requiring more tokens to encode sequences. In contrast, SentencePiece maintained more consistent token lengths between vocabulary and test data. WordPiece occupied a middle ground, offering a balance between efficiency and consistency. Notably, BPE showed superior performance in contextual exponence, consistently producing more contextually specialized tokens compared to its counterparts, particularly at larger vocabulary sizes. This suggests BPE's tokenization strategy may be more effective at capturing meaningful contextual relationships within protein data.

However, our analysis also revealed significant limitations. A key challenge across all tokenizers was their inability to consistently respect protein domain boundaries. Although BPE showed marginally better performance with smaller vocabularies, this general limitation suggests that current NLP-derived tokenizers require substantial adaptation to preserve structural units within proteins effectively.

The examination of linguistic laws provided further insights into the tokenizers' suitability for protein sequence analysis. While the tokenizers showed reasonable alignment with the brevity law, their partial compliance with Zipf's law and deviation from Menzerath's law suggest that protein sequences may follow distinct distribution patterns from natural language. Heap's law analysis demonstrated the tokenizers' ability to adjust vocabulary growth with sequence data size, though protein data exhibited greater variability compared to natural language text.

These deviations from established linguistic laws point to a fundamental difference between protein sequences and natural language, suggesting the possibility of unique biological linguistic laws governing protein structures. This observation opens new avenues for research, including investigating whether similar patterns exist in other biological systems and developing specialized tokenization approaches for protein sequences.

Our findings have important implications for the future development of protein sequence analysis tools. While current NLP tokenizers provide a valuable foundation, there is a clear need for domain-specific enhancements. Future tokenizers should be specifically designed to maintain domain boundary integrity, better capture protein-specific structural units, ensure consistent performance across diverse datasets, and balance contextual relevance with efficient segmentation.

In conclusion, this study advances our understanding of how NLP tokenization methods perform with protein sequences while highlighting the need for specialized approaches. The development of protein-specific tokenization strategies that address the identified limitations will be crucial for improving our ability to analyze and understand the complex language of proteins.

\section*{Acknowledgment}
We would like to thank Gökçe Uludoğan for her helpful comments on the manuscript.

This work is supported by ERC grant (LifeLU, 101089287). Views and opinions expressed are however those of the author(s) only and do not necessarily reflect those of the European Union or the European Research Council Executive Agency. Neither the European Union nor the granting authority can be held responsible for them.

\section*{Data Availability}
All data underlying this work, including source code, is available at \url{https://github.com/boun-tabi-lifelu/linguistics-meet-proteins}.

\bibliographystyle{ieeetr}
\bibliography{references}
\end{document}